\PassOptionsToPackage{numbers,sort&compress,square}{natbib}
\documentclass{article}
\usepackage{amssymb}
\usepackage{booktabs}
\usepackage{amssymb}
\usepackage{graphicx}
\usepackage{amsmath}
\usepackage{wrapfig}
\usepackage{caption}
\usepackage{color}
\usepackage{makecell}
\usepackage{pifont} 
\usepackage{tikz}   
\usepackage{bbm}

\usepackage{graphicx}
\usepackage{mwe} 
\usepackage{multirow}
\usepackage{makecell}  
\usepackage{xcolor}
\usepackage{pifont}
\definecolor{turquoise}{cmyk}{0.65,0,0.1,0.3}
\definecolor{purple}{rgb}{0.65,0,0.65}
\definecolor{dark_green}{rgb}{0, 0.5, 0}
\definecolor{orange}{rgb}{0.8, 0.6, 0.2}
\definecolor{red}{rgb}{0.8, 0.2, 0.2}
\definecolor{darkred}{rgb}{0.6, 0.1, 0.05}
\definecolor{blueish}{rgb}{0.0, 0.3, .6}
\definecolor{light_gray}{rgb}{0.7, 0.7, .7}
\definecolor{pink}{rgb}{1, 0, 1}
\definecolor{greyblue}{rgb}{0.25, 0.25, 1}
\definecolor{brinkpink}{rgb}{0.98, 0.38, 0.5}
\definecolor{codegreen}{rgb}{0,0.6,0}
\definecolor{codegray}{rgb}{0.5,0.5,0.5}
\definecolor{codepurple}{rgb}{0.58,0,0.82}
\definecolor{backcolour}{rgb}{0.95,0.95,0.92}

\usepackage[utf8]{inputenc}
\usepackage{tcolorbox}
\tcbuselibrary{breakable}
\tcbuselibrary{listings,skins}
\usepackage{listings}
\newcommand{\ourmethod}{\textsc{LodeStar}}
\newcommand{\ourmethodpc}{\textsc{LodeStar-PC}}
\newcommand{\ourmethodpose}{\textsc{LodeStar-Pose}}
\usepackage[final]{corl_2025} 

\title{LodeStar: Long-horizon Dexterity via Synthetic Data Augmentation from Human Demonstrations}

\author{%
  Weikang Wan\textsuperscript{1}\thanks{Equal contribution.}\quad
  Jiawei Fu\textsuperscript{1}\footnotemark[1]\quad
  Xiaodi Yuan\textsuperscript{1}\quad
  Yifeng Zhu\textsuperscript{2}\quad
  Hao Su\textsuperscript{1}\\[0.5ex]
  \normalfont
  \textsuperscript{1}University of California San Diego \quad
  \textsuperscript{2}The University of Texas at Austin \quad
}

\begin{document}
\maketitle



\vspace{-4mm}
\begin{abstract}
Developing robotic systems capable of robustly executing long-horizon manipulation tasks with human-level dexterity is challenging, as such tasks require both physical dexterity and seamless sequencing of manipulation skills while robustly handling environment variations.
While imitation learning offers a promising approach, acquiring comprehensive datasets is resource-intensive.
In this work, we propose a learning framework and system \ourmethod{} that automatically decomposes task demonstrations into semantically meaningful skills using off-the-shelf foundation models, and generates diverse synthetic demonstration datasets from a few human demos through reinforcement learning. 
These sim-augmented datasets enable robust skill training, with a Skill Routing Transformer policy effectively chaining the learned skills together to execute complex long-horizon manipulation tasks.
Experimental evaluations on three challenging real-world long-horizon dexterous manipulation tasks demonstrate that our approach significantly improves task performance and robustness compared to previous baselines. Videos are available
at \href{https://lodestar-robot.github.io/}{lodestar-robot.github.io}.
\end{abstract}

\keywords{Dexterous Manipulation, Imitation Learning,  Sim-to-Real}
\vspace{0mm}


\begin{figure}[!h]
    \centering
    \includegraphics[width=0.92\textwidth]{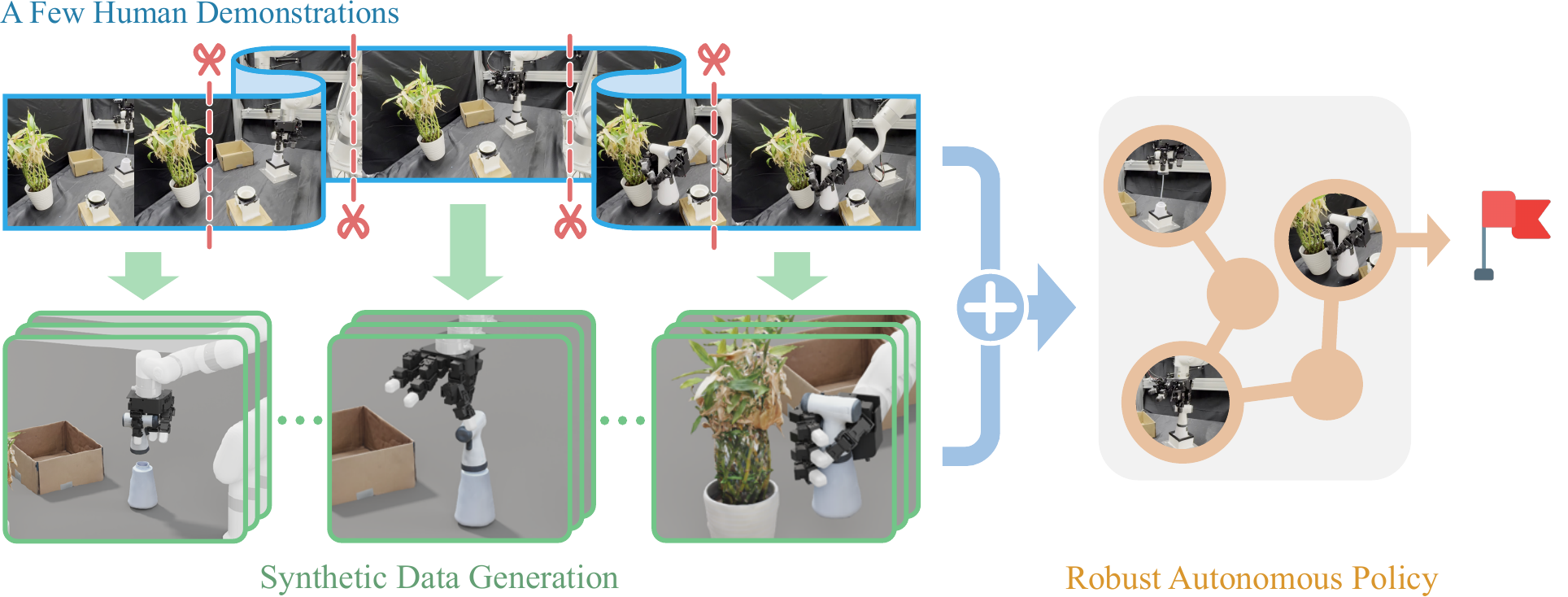}
    \caption{\textbf{We present \ourmethod{}, a framework for learning robust long-horizon dexterous manipulation from a few human demonstrations.} \ourmethod{} segments demonstrations into sequential skills using off-the-shelf foundation models, and augments each skill via simulation-based residual reinforcement learning. The synthesized skill data is co-trained with real-world data, and a Skill Routing Transformer composes the learned skills into a robust autonomous policy capable of completing complex real-world tasks.}
    \label{fig:pull-figure}
\end{figure}
\vspace{-2mm}

\section{Introduction}
\vspace{-2mm}
Developing multi-fingered robotic systems capable of long-horizon manipulation with human-level dexterity has been a longstanding goal in robotics research. 
Consider everyday activities such as watering a plant with a spray bottle as shown in Fig.~\ref{fig:pull-figure}: this requires not only physical dexterity to insert and twist the nozzle, lift and hold the bottle upright, and press the trigger, but also the ability to seamlessly sequence these distinct behaviors while robustly handling environment variations. Imitation learning (IL) from human demonstrations~\citep{brohan2022rt,jang2022bc,mandlekar2021matters} offers an effective pathway for teaching robots such behaviors. 
However, for such tasks, acquiring massive, comprehensive, and diverse datasets necessary to ensure policy robustness across potential perturbations and environmental variations encountered in the real world is often costly and resource-intensive~\citep{ravichandar2020recent,o2024open}. 
This challenge motivates our investigation into methods that leverage structured data generation from limited demonstrations, aiming to enable robots to perform long-horizon dexterous manipulation tasks robustly.

Prior works have explored generating large datasets from a few demonstrations using replay-based transformation with motion planning~\citep{mandlekar2023mimicgen,jiang2024dexmimicgen,garrett2024skillmimicgen,xue2025demogen}. 
While effective, these methods are often limited to parallel grippers or fixed-hand motions. Moreover, naive composing the adapted segments can lead to low-quality or low-diversity demonstrations, hindering generalizable policy learning.
Another direction leverages reinforcement learning (RL) in simulation to synthesize diverse data. Recent advances have enabled dexterous skills such as grasping ~\citep{xu2023unidexgrasp,lum2024dextrah,fang2025anydexgrasp} and in-hand reorientation~\citep{andrychowicz2020learning,qi2023hand,wang2024lessons}, as RL exploration can discover varied and robust behaviors.
However, these results are typically limited to single-stage tasks, rely on hand-tuned simulators and rewards, and remain sensitive to the sim-to-real gap.
In parallel, several works decompose long-horizon manipulation into sequential skills and chain them to solve complex tasks~\citep{konidaris2009skill,konidaris2012robot,clegg2018learning,bagaria2019option,lee2020learning,MCPPeng19}, reducing both the complexity of data generation and policy learning.
Yet, smooth and reliable skill transfer remains challenging due to state distribution mismatches at skill boundaries~\citep{gu2022multi}, often requiring additional regularization~\citep{lee2021adversarial,chen2024scar} or optimization~\citep{chen2023sequential} techniques, which may destabilize policy learning. 
Despite these advances, robust long-horizon dexterous manipulation from limited data still faces two key challenges: (1) how to automatically segment tasks into semantically meaningful skills with appropriate initial and terminal state distributions; and (2) how to generate diverse training data to robustify individual skills, and effectively chain them into a coherent policy that completes the full task.

To address these challenges, we propose a structured and scalable framework \ourmethod{} that decomposes long-horizon dexterous manipulation into three stages: skill segmentation, synthetic data generation for robust skill learning, and skill composition via a \textit{Skill Routing Transformer (SRT)} policy.
We first segment demonstrations into sequential manipulation skills and intermediate transition motions by leveraging foundation models (VFMs, VLMs)
~\citep{karaev2024cotracker,tang2023emergent,openai_o3_2025} to extract visual, spatial, and contact cues directly from raw videos, avoiding manual annotation or predefined primitive skills.
For each manipulation skill, we train a robust policy via residual reinforcement learning in simulation environments generated through real-to-sim techniques with augmentation and domain randomization. 
The learning process is grounded in a small number of real demonstrations, ensuring task relevance and avoiding reward engineering.
We further synthesize diverse and physically feasible transition trajectories between skills, eliminating the need for explicit goal specification or slow execution speed of real-world motion planning.
Finally, we train a SRT policy on the generated data to compose the learned skills, enabling coherent execution of the full long-horizon task.
Our key contribution is a scalable learning framework \ourmethod{} for robust long-horizon dexterous manipulation from a few human demos. By combining automatic segmentation via foundation models, real-to-sim synthetic data generation for skill acquisition, and skill composition for sim-to-real deployment, \ourmethod{} enables complex manipulation sequences without requiring extensive task-specific data or manual engineering effort. We validate its effectiveness on three challenging real-world tasks, showing that it outperforms SOTA baselines by an average of 25\% success rate.
\vspace{-2mm}
\section{Related Work}
\vspace{-2mm}
\subsection{Dexterous Manipulation}
\vspace{-2mm}
Dexterous manipulation remains a long-standing and significant challenge in robotics~\citep{salisbury1982articulated,mordatch2012contact,akkaya2019solving,helearning}, primarily due to the high degrees of freedom involved. 
Traditional planning and control approaches~\citep{fearing1986implementing,han1998dextrous,rus1999hand,bai2014dexterous} often rely on simplified system models, which struggle to scale to complex, long-horizon tasks.
Recent advances have demonstrated the effectiveness of learning-based methods for specific dexterous skills such as grasping~\citep{lum2024dextrah,xu2023unidexgrasp, wan2023unidexgrasp++,fang2025anydexgrasp}, in-hand reorientation~\citep{handa2023dextreme,qi2023hand,wang2024lessons}, and tool use~\citep{liu2022hoi4d,shaw2024bimanual,lin2024twisting, chen2024object,yin2025dexteritygen}.
However, these methods typically target isolated skills rather than integrated, sequential behaviors.
In contrast, we tackle the more challenging task of sequencing multiple dexterous skills~\citep{chen2023sequential}, which requires both fine-grained control and temporal coordination.

\subsection{Synthetic Data Generation for Robotic Manipulation}
\vspace{-2mm}
Synthetic simulation data have been used to alleviate the burden of real-world data collection and improve policy generalization in robotic manipulation.
Recent studies~\citep{bousmalis2018using,nasiriany2024robocasa,wang2024cyberdemo,wang2024poco,ankile2024imitation,maddukuri2025sim,wei2025empirical,xue2025demogen,geng2025roboverse,liunavid} show that co-training policy with simulation data can greatly improve policy performance and robustness by diversifying the training distribution beyond what real-world data alone can offer.
Real-to-sim approaches have been used to generate realistic and diverse simulation assets and scenes from real-world inputs via 3D reconstruction~\citep{henry2012rgb,tancik2023nerfstudio,jiang2022ditto,torne2024reconciling,patel2025real,ye2025video2policy,xia2025drawer}, inverse graphics~\citep{chen2024urdformer}, and foundation model-assisted generation~\citep{wang2023robogen,nasiriany2024robocasa,dai2024automated,mandi2024real2code}.
Techniques for sim-to-real transfer include domain randomization~\citep{tobin2017domain,peng2018sim,chebotar2019closing,mehta2020active,andrychowicz2020learning,handa2023dextreme,wang2024cyberdemo}, system identification~\citep{kumar2021rma,evans2022context,muratore2022neural,huang2023went,ren2023adaptsim,memmel2024asid} and simulator augmentation~\citep{bousmalis2018using,rao2020rl,ho2021retinagan}.
However, naive randomization and augmentation are insufficient to close the sim-to-real gap~\citep{dai2024automated,wang2024cyberdemo}, particularly in long-horizon tasks involving multi-stage and multi-object interactions, where the large variation space hinders effective learning.
To address this, our pipeline improves training robustness and efficiency by segmenting demonstrations into multiple stages and applying demonstration-guided augmentation and randomization to expand simulation coverage for each stage.

\subsection{Skill Chaining for Long-horizon Robotic Manipulation}
\vspace{-2mm}
Established methods tackle long-horizon tasks by decomposing them into simpler, reusable subtasks, reducing the agent's decision burden via temporally extended actions or skills~\citep{sutton1999between,konidaris2019necessity}.
Skill discovery approaches fall into three categories: predefined measures ~\citep{lee2019composing,kulkarni2016hierarchical,oh2017zero,zhu2021hierarchical,silver2022learning,nasiriany2022augmenting,garrett2024skillmimicgen,gu2022multi,mandlekar2023mimicgen,jiang2024dexmimicgen,cheng2023league,guo2025srsa}, skill discovery from demonstrations~\citep{pastor2009learning,konidaris2012robot,su2018learning,xu2018neural,kipf2019compile,zhu2022bottom,li2023dexdeform,xu2023xskill,wan2024lotus,wang2024skild}, and unsupervised self-exploration~\citep{schmidhuber1990towards,konidaris2009skill,vezhnevets2017feudal,bacon2017option,nachum2018data,bagaria2019option,eysenbach2018diversity,hausman2018learning}.
Naively sequential chaining skills often lead to “hand-off” failures, where the terminal state of one skill falls outside the feasible range of the succeeding skill.
Prior work mitigates this by updating each skill to encompass the terminal state of the preceding skill~\citep{konidaris2009skill,konidaris2012robot,clegg2018learning,bagaria2019option,lee2020learning,MCPPeng19}, or by aligning skill boundaries through reward shaping via adversarial~\citep{lee2021adversarial,chen2024scar} or bidirectional training~\citep{chen2023sequential}.
Closest to our work,~\citep{lee2019composing} learns a transition policy to bridge skills, but assumes it can be trained via random exploration and is limited to simple, predefined primitives.
However, these methods typically require either extensive environment interaction~\citep{kulkarni2016hierarchical,lee2019composing,nasiriany2022augmenting,chen2023sequential,chen2024scar}, annotated demonstrations~\citep{mandlekar2023mimicgen,garrett2024skillmimicgen}, or high computational resources~\citep{mishra2023generative,mishragenerative}, limiting their scalability to contact-rich, high-DoF dexterous tasks.
Our work addresses this by segmenting skills from a small number of demonstrations, and combines insights from skill chaining with recent advances in foundation models~\citep{karaev2024cotracker,tang2023emergent,haldar2025point}.
\vspace{-1mm}
\section{Problem Formulation}
\label{problem_formulation}
\vspace{-2mm}
We formulate our learning problem as a Markov Decision Process (MDP) defined by the tuple $(\mathcal{S}, \mathcal{A}, R, P, \rho_0, \gamma)$, where $\mathcal{S}$ is the state space, $\mathcal{A}$ is the action space, $R(s, a, s')$ is the reward function, $P(s' \mid s, a)$ is the transition distribution, $\rho_0$ is the initial state distribution, and $\gamma$ is the discount factor.
We aim to solve a long-horizon task $\mathcal{T}$, which can potentially be decomposed into a sequence of subtasks $\{\mathcal{T}_1, \mathcal{T}_2, \ldots, \mathcal{T}_K\}$, where $K$ is the total number of subtasks and varies for each long-horizon task. Each subtask $\mathcal{T}_i$ is associated with an initiation set $\mathcal{I}_i \subset \mathcal{S}$ and a termination set $\mathcal{E}_i \subset \mathcal{S}$, indicating the valid start and end states for completing subtask $\mathcal{T}_i$. A policy $\pi: \mathcal{S} \rightarrow \mathcal{A}$ induces a distribution over trajectories, and is optimized to maximize the expected sum of discounted rewards $\mathbb{E}_{\pi} \left[ \sum_{t=0}^{T-1} \gamma^t R(s_t, a_t, s_{t+1}) \right]$, where $T$ is the episode horizon.
We assume access to a small set of full expert demonstrations $\mathcal{D}^e = \{\tau^e_j\}_{j=1}^N$, where $\tau^e_j$ is a trajectory of state-action pairs.


\begin{figure}[t]
    \centering
    \includegraphics[width=1.0\textwidth]{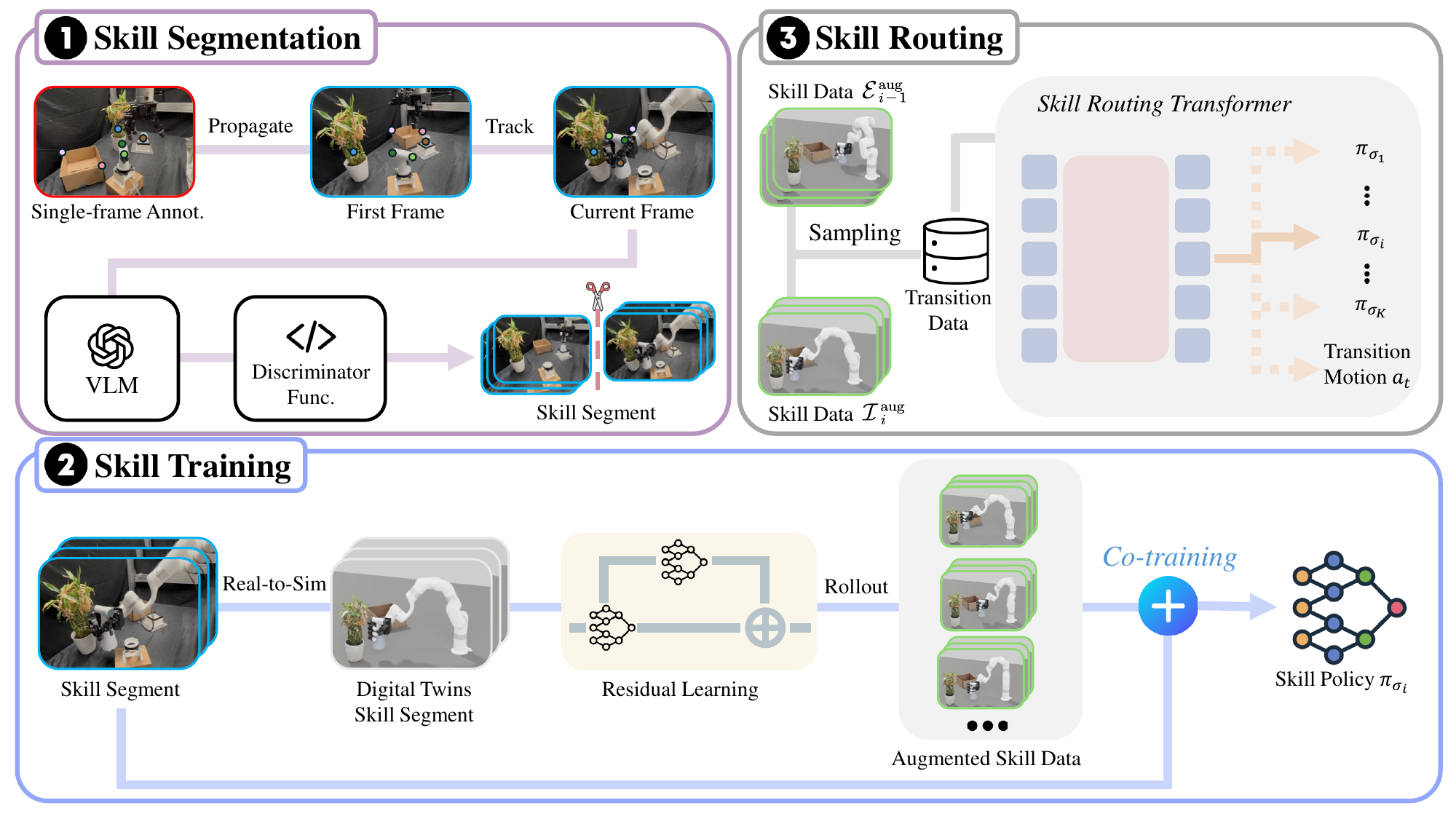}
    \caption{\textbf{\ourmethod{} Pipeline. } \ourmethod{} consists of three sequential stages. \textbf{(1)} Human demonstrations are segmented into manipulation and transition phases using discriminators built from VLM. \textbf{(2)} Each skill is augmented via residual RL in simulation to train robust policies.\textbf{(3)} Skill Routing Transformer policy composes and selects skills or transition actions during execution, enabling coherent long-horizon manipulation in real-world settings.}
    \label{fig:pipeline}
    \vspace{-4mm}
\end{figure}

\section{Approach}
\vspace{-2mm}

We introduce \ourmethod{}, a framework designed to enable robust long-horizon dexterous manipulation by leveraging a few human demonstrations and sim-augmented data. An overview of the system architecture is shown in Fig.~\ref{fig:pipeline}.
\ourmethod{} comprises three key components:
\textbf{(1) Skill Segmentation (Sec.~\ref{sec:skill_segmentation}):} decomposing human demonstrations into motion and manipulation phases using per-frame skill discriminators derived from vision-language models and tracked keypoints;
\textbf{(2) Synthetic Data Generation for Robust Skill Policies Learning (Sec.~\ref{sec:sub_policy_learning}):} training individual robust skill policies enhanced by synthetic demonstrations from residual reinforcement learning (RL) policy;
\textbf{(3) Skill Composition via Skill Routing Transformer (SRT) policy (Sec.~\ref{sec:transfer_policy}):} learning a SRT policy to sequence and compose the learned skills to accomplish long-horizon tasks.

\vspace{-3mm}
\subsection{Skill Segmentation}
\label{sec:skill_segmentation}
\vspace{-2mm}
In long-horizon dexterous manipulation, human behavior often exhibits a natural decomposition which alternates between two modes: (i) \textit{transition} motions corresponding to inter-skill transit movements~\citep{garrett2024skillmimicgen,sundaresan2024s} that involve either no object interaction or only passive contact, and (ii) \textit{manipulation skill} involving fine-grained, contact-rich manipulations primarily actuated by the hand.
We propose decomposing each expert demonstration $\tau^e_j = \{(s_t, a_t)\}_{t=1}^T$ into an alternating sequence of manipulation skill and transition motion segments:
$\tau^e = \tau_{\sigma_1} \rightarrow \tau_{\mu_1} \rightarrow \tau_{\sigma_2} \rightarrow \cdots \rightarrow \tau_{\mu_{K-1}} \rightarrow \tau_{\sigma_K}$,
where each $\tau_{\sigma_i}$ denotes the demonstration of a semantically meaningful manipulation skill $\sigma_i$ (e.g., inserting, twisting) to solve the subtask $\mathcal{T}_i$, and each $\tau_{\mu_i}$ denotes a motion segment serving as a transition between skills.
To identify skill segments, we propose a pipeline that leverages vision foundation models and vision-language models to construct a per-frame \textit{discriminator function} for each skill. Each discriminator $d_i(s_t): \mathcal{S}\rightarrow \{0, 1\}$ determines whether a given frame belongs to the skill $\sigma_i$, and is formulated as a conjunction of two constraints: $ d_i(s_t) = \mathbbm{1} \left[ C^{\text{point}}_i(s_t) \wedge C^{\text{contact}}_i(s_t) \right]$, where $C^{\text{point}}_i(\cdot)$ is a keypoint spatial relation constraint and $C^{\text{contact}}_i(\cdot)$ is a fingertip contact constraint which describes whether the fingertip has contact with the object of interest (see Appendix for more details).
Unlike prior works~\citep{zhang2024universal,huang2024rekep} that define a skill by its terminal state distribution $\mathcal{E}_i$, we classify skills at the frame level.  This formulation is more tractable in dexterous skills, where end states are typically task-specific and diverse across human demonstrations~\citep{chen2023sequential}, whereas the execution process exhibits consistent spatial and contact patterns.

\textbf{Keypoint Proposal.}
To obtain skill-centric spatial constraints, we annotate $N$ semantically meaningful keypoints $\{\hat{p}_l\}_{l=1}^N$ on task-relevant objects in the first frame of a random reference demonstration for each task. These annotated keypoints serve as priors for other demonstrations.
We then use the semantic correspondence model DIFT~\citep{tang2023emergent} to propagate the initial keypoints to the first frame of other demonstrations, obtaining consistent keypoints across demonstrations. 
Then, we apply Co-Tracker~\citep{karaev2024cotracker} to track the keypoints $\{p_l^{(j)}(t)\}_{t=1}^T$ across each trajectory.
This procedure yields per-frame object keypoints, with only a single manual annotation required.

\textbf{Stage-wise Discriminator Function Generation.}
Given the language instruction of the task, we use visual prompting with OpenAI o3 model~\citep{openai_o3_2025} to generate the number of semantic manipulation skill stages and corresponding Python functions for each stage-wise discriminator $d_i^{(s)}(s_t)$. 
Following prior works~\citep{huang2023voxposer,huang2024rekep}, we leverage VLMs to specify structured relations (e.g., relative distances, alignment conditions) as symbolic arithmetic expressions, rather than manipulating numerical values directly. This approach improves the generalizability and interpretability of the discriminators.

\vspace{-2mm}
\subsection{Synthetic Data Generation for Robust Skill Policies Learning}
\label{sec:sub_policy_learning}
\vspace{-2mm}
To improve the robustness and generalization of each skill policy, we leverage simulation to train RL policies under diverse and augmented conditions. 
For each skill $\sigma_i$, we construct a realistic simulation environment, augment its initial and terminal state distributions as well as physical parameters, transfer the corresponding real-world segments into simulation by state estimation, train a base imitation learning policy $\pi_{\sigma_i}^{\text{base}}$ using the transferred segment data, and train a residual RL policy $\pi_{\sigma_i}^{\text{res}}$ complementing the base policy.
We then collect simulated demonstrations under real-world observation space and co-train the final policy $\pi_{\sigma_i}$ using both real and simulated data.


\noindent \textbf{Real-to-Sim Transfer.}
For each skill $\sigma_i$, we construct a simulation environment that closely matches the real-world scene.
To achieve this, we first use existing off-the-shelf 3D reconstruction models~\citep{arcode2022} to generate textured object meshes from multi-view images. For articulated objects, following prior work~\citep{torne2024reconciling},  we manually separate each mesh into individual links and define their kinematic relationships by adding articulations.
To model physical parameters (e.g., mass, friction), we avoid explicit system identification, which typically requires extensive real-world interaction~\citep{bohg2017interactive,pfaff2025scalable}. Instead, we adopt domain randomization: during training, dynamics parameters are sampled from a predefined range, which is effective for sim-to-real transfer.
From expert demonstrations  $\tau_{\sigma_i}$, we extract the observed initial and terminal states for each skill $\sigma_i$, denoted as $\mathcal{I}_i^{\text{real}}$ and $\mathcal{E}_i^{\text{real}}$, respectively. These correspond to the observed states at the boundaries of skill segments. To improve robustness, we augment both sets via object-centric perturbations (e.g., pose noise, small translations, and rotations. See Appendix for details), yielding:
$\mathcal{I}_i^{\text{aug}} = \text{Augment}(\mathcal{I}_i^{\text{real}}), \mathcal{E}_i^{\text{aug}} = \text{Augment}(\mathcal{E}_i^{\text{real}})$.
We use FoundationPose~\citep{wen2024foundationpose} to estimate the 6D poses of the objects involved for real-to-sim demonstration transfer.

\noindent \textbf{Skill Policies Training.}
Given the simulation environment for each skill $\sigma_i$ and converted demonstration segments from $\tau_{\sigma_i}$, the goal is to train a robust policy $\pi_{\sigma_i}$  within the corresponding initial distribution $\mathcal{I}_i^{\text{aug}}$ and end distribution $\mathcal{E}_i^{\text{aug}}$.
Specifically, we first use the converted demonstration segments to train a base policy $\pi_{\sigma_i}^{\text{base}}$ with Behavior Cloning (BC), i.e., $\pi_{\sigma_i}^{\text{base}} = \arg\max_{\pi_{\sigma_i}^{\text{base}}} \mathbb{E}_{(s_t, a_t)} \left[ \log \pi_{\text{base}}(a_t \mid s_t) \right]$, and then train a residual policy $\pi_{\sigma_i}^{\text{res}}$ using PPO~\citep{schulman2017proximal} with privileged state information and binary sparse reward. 
During RL training, the actions from both policies are combined $\pi_{\sigma_i}^{\text{base}}(s_t)$ + $\pi_{\sigma_i}^{\text{res}}(s_t)$.
We use orthogonal initialization~\citep{saxe2013exact} for the residual policy network and progressive exploration schedule~\citep{yuan2024policy} to stabilize training (see Appendix for more details).
After that, we rollout the trained policy initialized from $\mathcal{I}_i^{\text{aug}}$ to collect successful trajectories (end within $\mathcal{E}_i^{\text{aug}}$) with real-world observable state information in simulation. 
Finally, we co-train a policy $\pi_{\sigma_i}$ with the simulation trajectories and real-world trajectories using the same BC loss.
Notably, real-world data is used to both enable base policy training via state estimation and co-train with simulated data for enhanced robustness.

\vspace{-2mm}
\subsection{Skill Composition via Skill Routing Transformer (SRT) Policy}
\label{sec:transfer_policy}
\vspace{-2mm}
Given the trained skill policies $\pi_{\sigma_i}$, we aim to chain them together to achieve long-horizon manipulation.
To transfer from the terminal state of one skill to the initial state of the succeeding skill, a direct solution is to use motion planning in the real world. 
However, such method faces challenges in selecting reachable goals between the state distributions $\mathcal{E}_{i-1}^{\text{aug}}$ and $\mathcal{I}_{i}^{\text{aug}}$ which may require additional human-designed rules, and its computational expense often limits execution speed, especially in cluttered dexterous manipulation scenarios.
Instead, we generate diverse and physically plausible trajectories in simulation.
During data generation, we randomly sample state pairs $(s^{\text{end}}, s^{\text{start}}) \sim \mathcal{E}_{i-1}^{\text{aug}} \times \mathcal{I}_i^{\text{aug}}$, and generate smooth trajectories between them via motion planning. We then filter out the infeasible ones to ensure all trajectories reach targets without collision. This yields a transition dataset $\mathcal{D}^{\text{trans}}$ containing tuples $(s_t, a_t, k_t)$, where $k_t \in \{\texttt{transition}, \texttt{skill}_i\}$ denotes the execution stage.


We then use this dataset to train a transformer-based policy \textit{Skill Routing Transformer (SRT)} policy, which predicts the current transition motion and the stage (select either transition or one skill) for the next timestep.
SRT policy $\pi^{\text{g}}: \mathcal{O} \rightarrow \mathcal{A} \times \mathcal{K}$ maps the current observation $o_t$ to a dense action $a_t \in \mathcal{A}$ and a discrete stage $k_t \in \mathcal{K}$, where $\mathcal{K} = \{\texttt{transition}\} \cup \{\texttt{skill}_i\}_{i=1}^K$. In \texttt{transition} stage, the policy outputs low-level actions to bridge between two skills. In $\texttt{skill}_i$ stage, it executes the trained skill policy $\pi_{\sigma_i}$.
We implement $\pi^{\text{g}}$ using a Transformer-based architecture over a history of past observations, enabling temporal context awareness in mode prediction and action generation.
\begin{figure}[t]
    \centering
    \includegraphics[width= 1.0\linewidth]{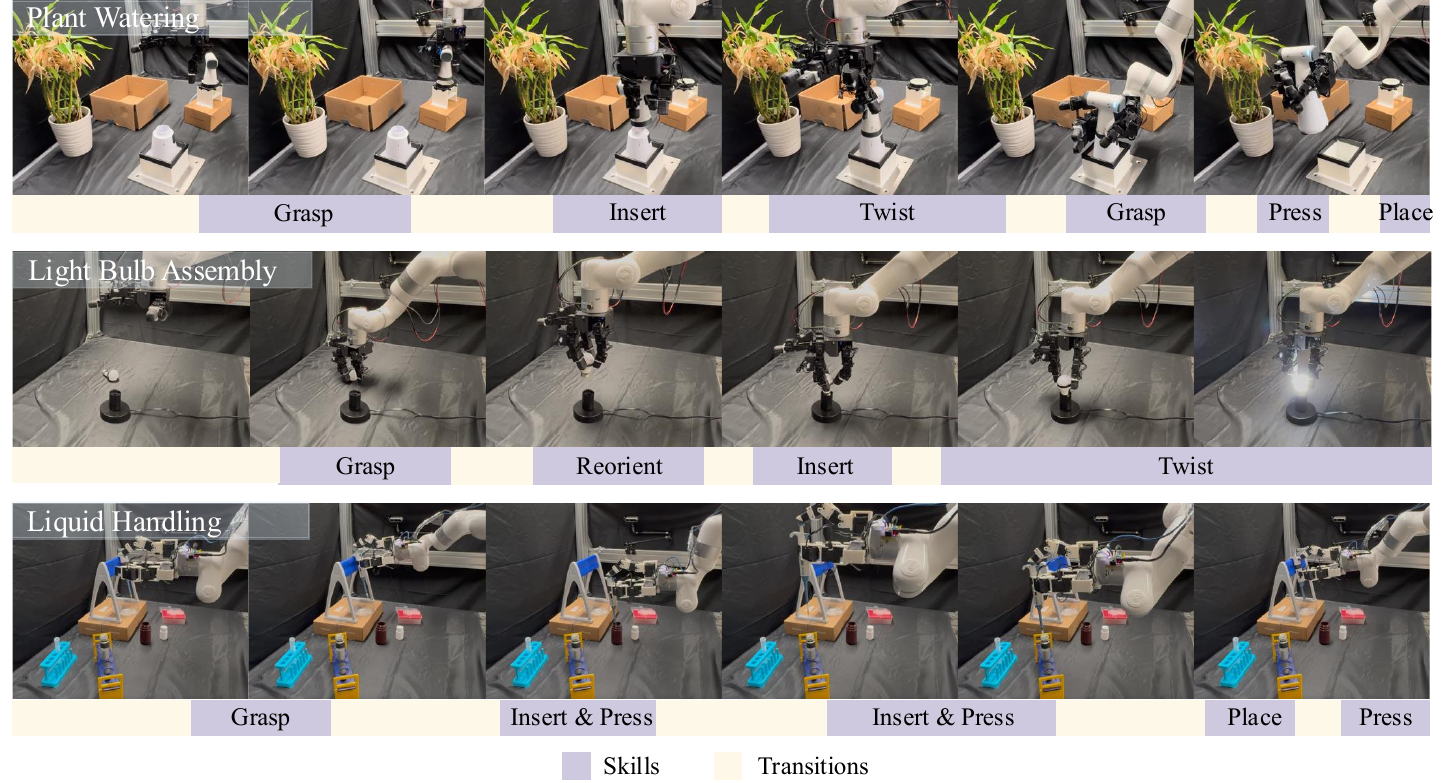}
    \caption{\textbf{Real-world Rollout Visualization.}We evaluate \ourmethod{} on three challenging real-world tasks (Plant Watering, Light Bulb Assembly, and Liquid Handling). The figure visualizes the segmentation results produced by our skill segmentation pipeline on human demonstration trajectories. Skill names are automatically generated by a vision-language model.}
    \label{fig:real_vis}
    \vspace{-6mm}
\end{figure}
\vspace{-4mm}

\section{Experimental Evaluation}
\vspace{-2mm}
We answer the following research questions through experiments:
\textbf{Q1.} Does \ourmethod{} lead to better transfer performance and robustness using the generated synthetic data?
\textbf{Q2.} Does \ourmethod{} help chaining multiple skill policies to accomplish long-horizon tasks?
\textbf{Q3.} Can \ourmethod{} help to improve policy generalization under out-of-distribution (OOD) conditions?  

\begin{figure}[t]
    \centering
    \includegraphics[width=1.0\textwidth]{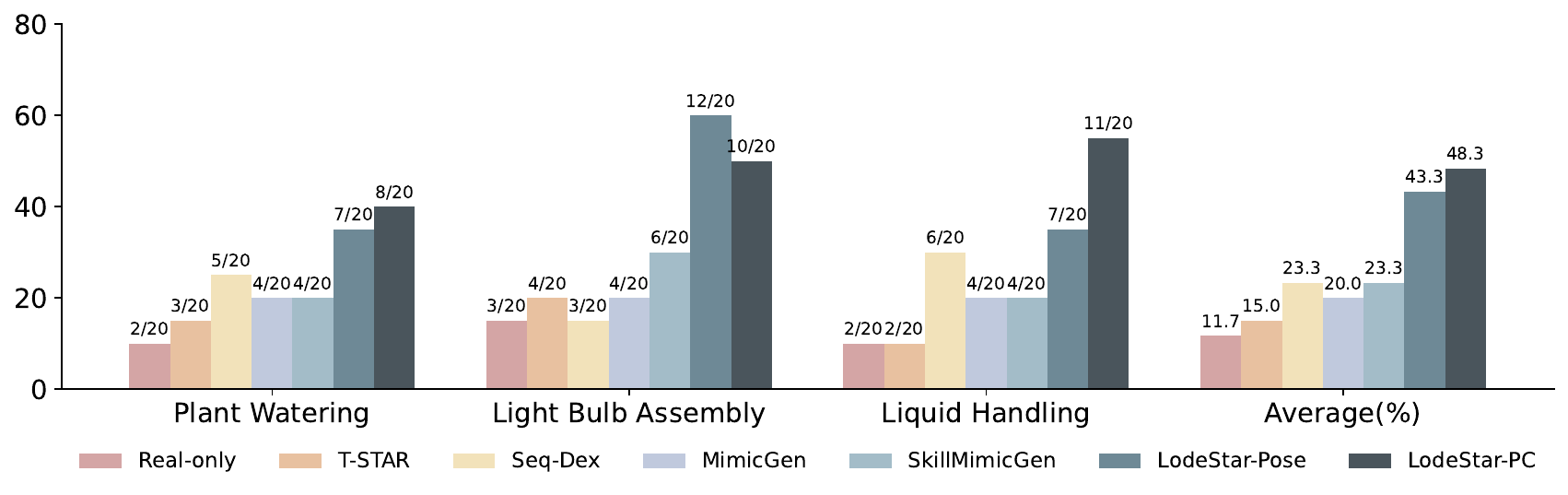}
    \caption{\textbf{Success rates across three challenging real-world tasks and their average.}
    \ourmethodpc{} demonstrates superior performance on average and across tasks, boosting average success by at least 25\% compared to the baselines.
    }
    \label{fig:exp_main}
    \vspace{-4mm}
\end{figure}

\noindent \textbf{Experimental Setup}
We evaluate our framework on 3 challenging real-world dexterous manipulation tasks requiring both fine-grained precision and reactivity (as shown in Fig.~\ref{fig:real_vis}):
1) \textbf{Liquid Handling} 
involves the robot picking up a pipette from a rack, aspirating liquid from a reagent bottle, dispensing it into a test tube, returning the pipette to the rack, and disposing of the used tip.
2) \textbf{Plant Watering}
requires grasping a spray nozzle, positioning it onto a bottle, securely twisting it in place, and triggering it to water the plants while holding the bottle.
3) \textbf{Light Bulb Assembly}
tasks the robot with gripping a light bulb, reorienting the bulb in-hand for insertion, and precisely screwing the bulb into a base until lighting.
Our real-world setup uses an xArm7 robot paired with either a three-finger hand or a four-finger LEAP hand~\citep{shaw2023leaphand}, and utilizes 2 RealSense D435 cameras. Real world data is collected via human teleoperation using a Rokoko Glove to track finger movements and a VIVE Ultimate Tracker to track wrist pose. We use Isaacgym~\citep{makoviychuk2021isaac} for the simulation environments.
See Appendix for the detailed environment setup, system configuration and evaluation protocol.

\begin{wrapfigure}[16]{r}{0.45\textwidth}
    \vspace{-\baselineskip}
    \centering
    \includegraphics[width=\linewidth]{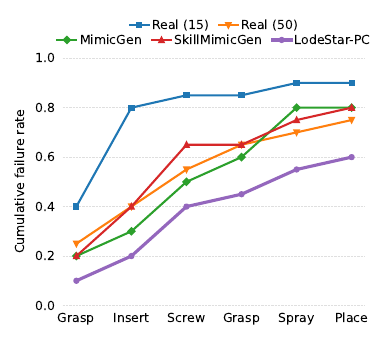}
    \caption{\textbf{Cumulative Failure Rate} on the Plant Watering task.}
    \label{fig:failure_curve}
\end{wrapfigure}
\vspace{-0\baselineskip}

\noindent \textbf{Baselines and Metrics}
We compare \ourmethod{} with the following baselines:
1) \textbf{Real-only} uses the same architecture and training as \ourmethod{}, but is only trained on real-world demonstrations.
2) Skill-chaining baselines: 
\textbf{T-STAR}~\citep{lee2021adversarial} uses terminal-state regularization, and \textbf{Seq-Dex}~\citep{chen2023sequential} uses bidirectional optimization. To ensure fairness, we use both simulation and real-world data for these methods. 
3) Automatic Data-Generation methods: \textbf{MimicGen}~\citep{mandlekar2023mimicgen} employs replay-augmentation on object-centric segments, and \textbf{SkillMimicGen}~\citep{garrett2024skillmimicgen} extends this idea to skill-level augmentation with skill chaining via motion planning.
We evaluate two variants of our framework, ``\ourmethodpose{}" conditions skill policy on estimated initial object poses (as real-time pose estimation is impractical due to its inference overhead), and  ``\ourmethodpc{}" (default option) conditions each skill policy on raw point-cloud observations.
Each of the methods evaluates 20 trials.
See Appendix for implementation details.

\subsection{Results and Analysis}
\noindent \textbf{\ourmethod{} is effective for real-to-sim-to-real transfer performance and robustness (Q1).}
As illustrated in Fig.~\ref{fig:exp_main}, \ourmethodpc{} achieves the best performance on average and in the three challenging tasks.
Compared to the best 
automatic data-generation 
baseline SkillMimicGen that uses replay-based demonstration transformation on the object-centric skill segments, \ourmethodpc{} boosts the average performance by 25\%. 
We attribute the improvement to two main factors:
1) \ourmethod{} adopts domain randomization in real-to-sim transfer to learn more robust sim-to-real behavior, while replay-based methods may fail due to the gaps from sensors and the controller in the bidirectional transfer between simulation and reality.
2) \ourmethod{} uses residual RL on a base policy from imitation learning to correct imperfect demonstrations and generalize across broader state distributions through exploration.
On the contrary, replay-based methods may fail to cover a diverse range of initial and terminal skill states because of the kinematic constraints in the source demonstrations.
Notably, \ourmethodpose{}, despite its simplicity—conditioning skill policy solely on the estimated initial object pose—performs effectively and even surpasses \ourmethodpc{} in the light bulb assembly task, which involves a less cluttered environment than the other two tasks. This suggests that when occlusion is minimal and pose estimation is reliable, \ourmethodpose{} serves as a simple yet effective solution.


\noindent \textbf{\ourmethod{} is effective for long-horizon skill chaining (Q2).}
As shown in Fig.~\ref{fig:exp_main}, \ourmethod{} significantly outperforms prior skill chaining methods. 
Compared with T-Star and Seq-Dex, which apply uni- or bi-directional optimization on skill boundaries, \ourmethod{} avoids such regularization for training stabilization and instead employs a more efficient stage transition to chain skills, boosting performance by over 25\%.
In addition, we observe that online motion planning in SkillMimicGen often fails because of state estimation error from occlusion and sensor perception errors, whereas \ourmethod{} chains skills trained on generated successful synthetic transitions.

\begin{wraptable}[9]{r}{0.46\textwidth}
    \vspace{-1\baselineskip}
    \centering
    \footnotesize
    \begin{tabular}{@{}lcc@{}}
        \toprule
        \textbf{Method (\# Demos)}& \textbf{Larger Init} & \textbf{Disturbances} \\
        & \textbf{Distribution} &  \\
        \midrule
        Real-only (15) & 0/20 & 1/20 \\
        Real-only (50) & 4/20 & 5/20 \\
        SkillMimicGen (15) & 5/20 & 4/20 \\
        \ourmethod{}(15) & \textbf{10}/20 & \textbf{8}/20 \\
        \bottomrule
    \end{tabular}
    \caption{\textbf{Success rates under OOD conditions} on the Light Bulb Assembly task.}
    \label{tab:ood}
\end{wraptable}
\noindent \textbf{\ourmethod{} is effective for improving policy robustness and generalization under OOD situations (Q3).} 
We evaluate policies by initializing with a larger distribution or applying disturbances on the Light Bulb Assembly task. 
As shown in Tab.~\ref{tab:ood}, although more real-world data (15 demos vs. 50 demos) boosts the performance from 0 to 20\% and 5\% to 25\%, \ourmethod{} achieves \ensuremath{\sim}2 times higher success rate with only 15 demos and exhibits emergent behavior under disturbances, thereby demonstrating superior robustness and generalization. See Appendix for more details and visualization.

\noindent \textbf{Cumulative Failure Rate Analysis. }
Fig.~\ref{fig:failure_curve} shows the cumulative failure rates throughout the task execution on the Plant Watering task. 
\ourmethodpc{} consistently achieves the lowest failure rate at each stage, demonstrating superior robustness and compounding stability. 
In contrast, methods trained purely on real-world demonstrations suffer from early failure accumulation, particularly after the ``Insert" and ``Screw" stage.
Both MimicGen and SkillMimicGen reduce early-stage failures but plateau in later stages.
These results underscore the effectiveness of our approach in mitigating error accumulation and ensuring reliable execution over extended task horizons.

\begin{wraptable}[10]{r}{0.46\textwidth}
    \vspace{-1\baselineskip}
    \centering
    \footnotesize
    \begin{tabular}{@{}lcc@{}}
        \toprule
        & \textbf{Simulation} & \textbf{Real-world} \\
        \midrule
        \ourmethod{} & 74\% & \textbf{10/20} \\
        w/ predefined skills & 65\% & 6/20 \\
        RL fine-tuning & 45\% & 3/20 \\
        w/o transition stage & 62\% & 6/20 \\
        w/o sim augmentation & 91\% & 4/20 \\
        w/o real co-training & 76\% & 7/20 \\
        \bottomrule
    \end{tabular}
    \caption{\textbf{Success Rates with Ablated Components} on the Light Bulb Assembly task.}
    \label{tab:ablation}
\end{wraptable}
\noindent \textbf{Ablations.}
We further evaluate the contribution of key components in \ourmethod{} through ablation studies on the Light Bulb Assembly task, as shown in Tab.~\ref{tab:ablation}.
Compared to using human pre-defined skills, our skill segmentation pipeline improves real-world success rate by 20\%, highlighting its effectiveness in capturing meaningful skill boundaries.
\ourmethod{} also significantly outperforms using RL fine-tuning (by 35\% in real world), benefiting from residual RL which stabilizes simulation training.
Removing the transition stage or simulation augmentation leads to 20\% and 30\% drops in real-world success respectively, indicating the importance of smooth skill composition and simulation distribution augmentation.
Finally, co-training with real-world data yields an additional 15\% improvement over simulation-only training, demonstrating its value in enhancing sim-trained policies.
These results again highlight the effectiveness of these key designs in \ourmethod{}.

\vspace{-2mm}
\section{Conclusion}
\vspace{-4mm}
To summarize, we presents \ourmethod{}, a system for automatic skill decomposition and synthetic data generation from human demonstrations for long-horizon dexterous manipulation tasks.
The sim-augmented datasets enable robust skill acquisition, while a global policy integrates and chains these skills into long-horizon executions, enhancing robustness and generalization in real-world environments.
The results on three real-world challenging tasks show that \ourmethod{} achieves 2 times higher success rate compared to the best baseline, while also generalizing to a broader range of environment variations. 
These highlight the potential of combining structured task representations with scalable synthetic data augmentation for efficient and generalizable dexterous robot learning.

\section{Limitations.}
Despite compelling results, our approach has several limitations, which also highlight multiple avenues for future works:
First, while our method demonstrates significant improvements over baseline approaches, the overall task performance has not yet reached saturation and exhibits room for further enhancement. Future research could investigate techniques like human-in-the-loop online correction mechanisms to potentially boost performance levels.
Second, the depth sensors used for constructing 3D point clouds, keypoint projection, and pose estimation can be unreliable—particularly when dealing with transparent or highly reflective objects (in our experiments, we manually applied opaque tape to transparent test tubes). Future work might integrate additional sensing modalities, such as tactile contact information, to improve both skill decomposition and policy learning.
Third, our real-to-sim pipeline currently overlooks dynamic parameters. Future efforts could apply system-identification techniques to model dynamics in simulation more precisely, thereby enhancing the fidelity of sim-to-real transfer.
Lastly, all experiments have been conducted using rigid objects. We did not investigate tasks involving deformable objects, as these objects remain challenging to reconstruct and simulate accurately.  
Extending our framework with advanced simulation and representation methods for deformable objects would broaden its applicability.

\clearpage


\bibliography{reference}  

\clearpage
{\large
\textbf{Supplementary Material}
}

\appendix
\section{Real Robot Platform}

In this section, we provide details about our real robot platform, including the hardware setup, teleoperation system, and environment setup.

\subsection{Hardware Setup}

\begin{figure}[htbp]
    \centering
    \includegraphics[width=1.0\textwidth]{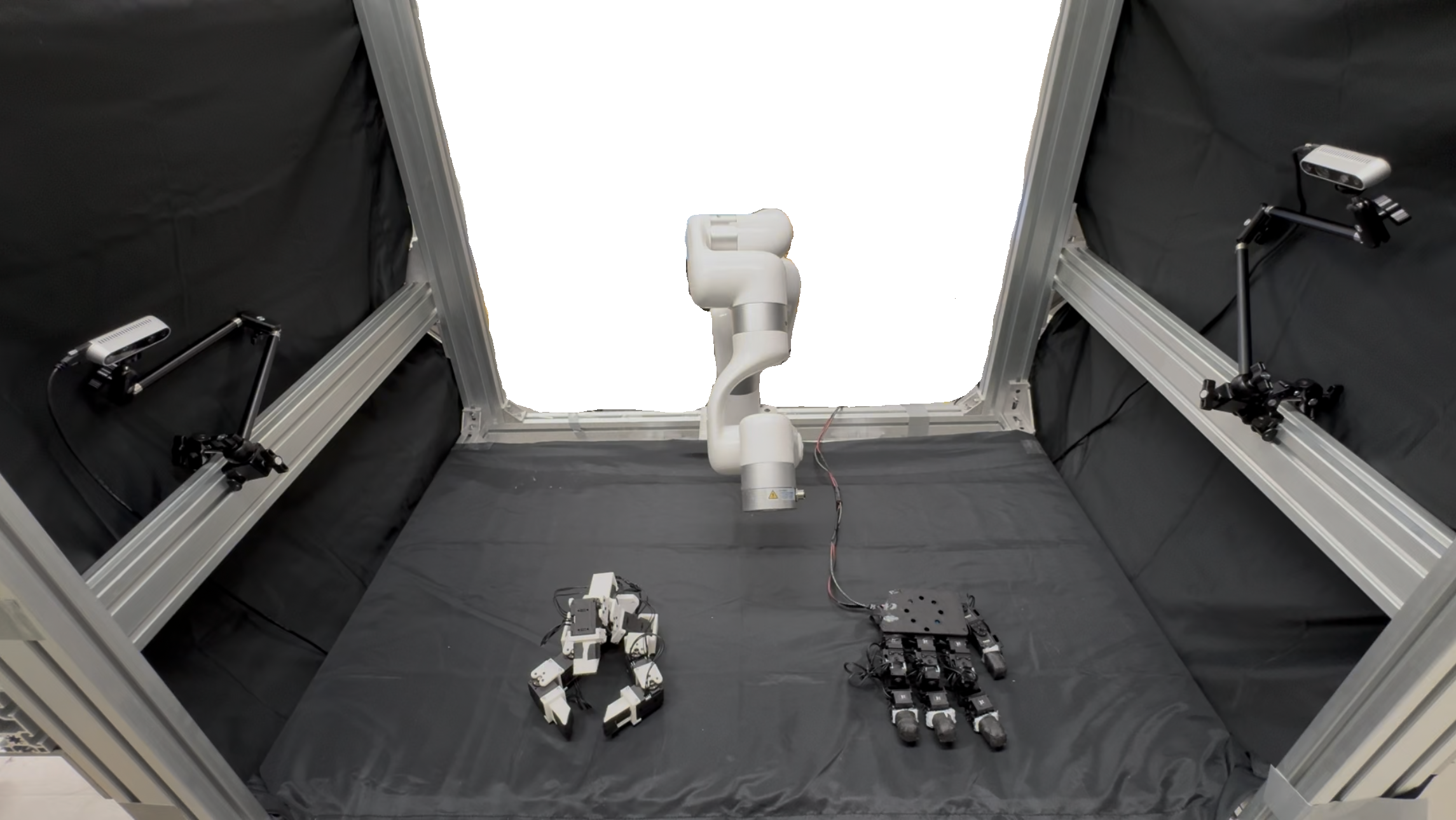}
    \caption{\textbf{Hardware Setup for \ourmethod{}}. 
    The hardware setup uses an xArm7 robot arm with two RealSense D435 cameras. We use two different dexterous robotic hands: \textbf{(Left)} a three-fingered hand with 9 degrees of freedom (DoF), and \textbf{Right} a four-fingered Leap hand with 16 DoF.}
    \label{fig:hardware-setup}
    \vspace{4mm}
\end{figure}

As shown in Fig.~\ref{fig:hardware-setup}, our hardware system consists of an xArm7 robot mounted on the tabletop.
We use either a three-finger hand or a four-finger LEAP hand~\citep{shaw2023leaphand} at the end-effector of the xArm7 robot.
There are two RealSense D435 cameras mounted on the left and right sides of the workspace, respectively.

\subsubsection{Robot Arm}

We use a joint velocity controller from BunnyVisionPro~\citep{bunny-visionpro} to control the xArm7 robot.
The controller receives joint position commands at 20 Hz. These commanded positions are used by a Proportional-Derivative (PD) controller to compute corresponding joint velocities. For safety, these joint velocities are subsequently clipped before being used to command the robot's motors at a rate of 250 Hz.

\subsubsection{Dexterous Hand}

\begin{figure}[htbp] 
    \centering 
        \includegraphics[width=0.48\textwidth]{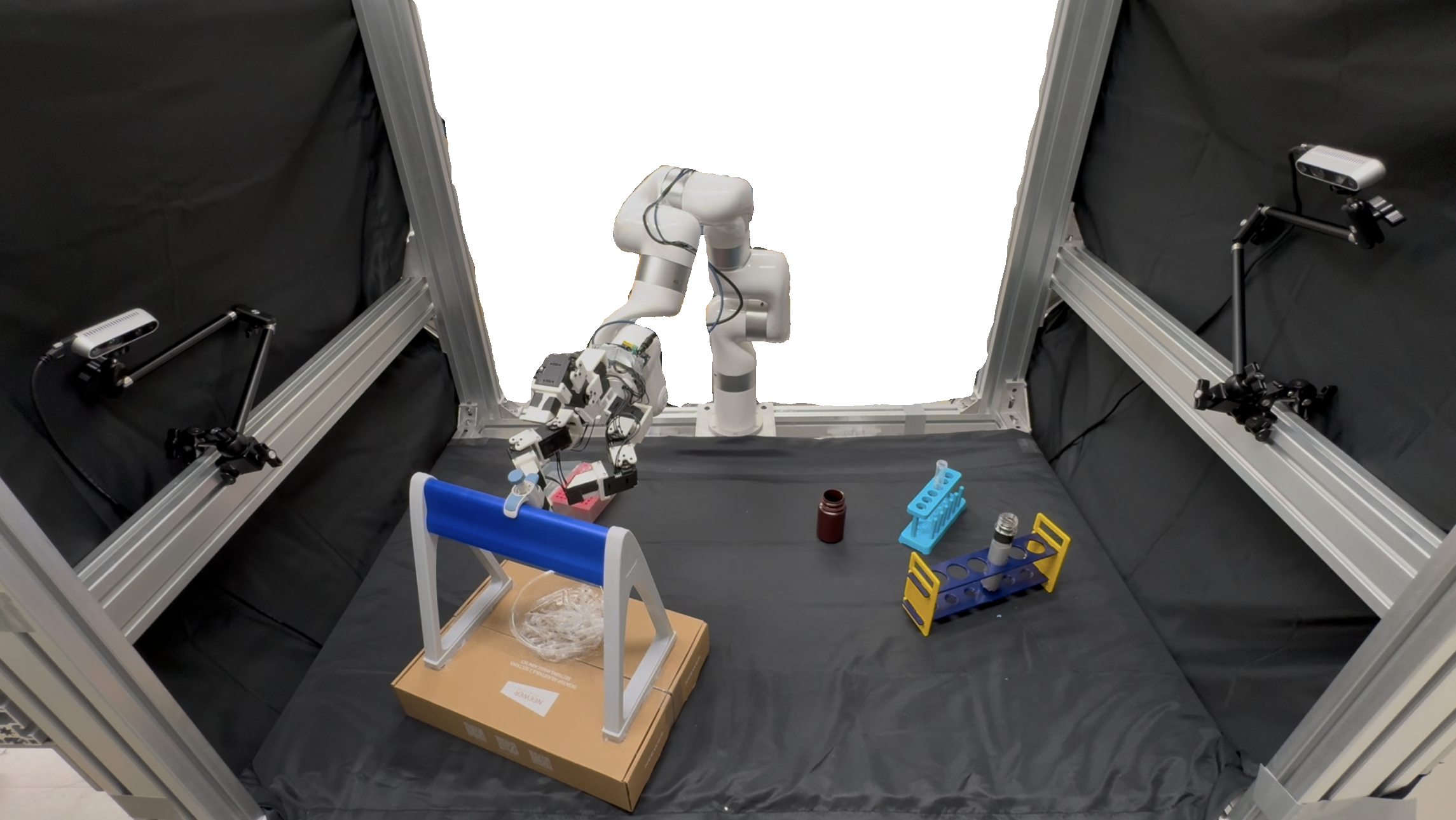} 
        \includegraphics[width=0.36\textwidth]{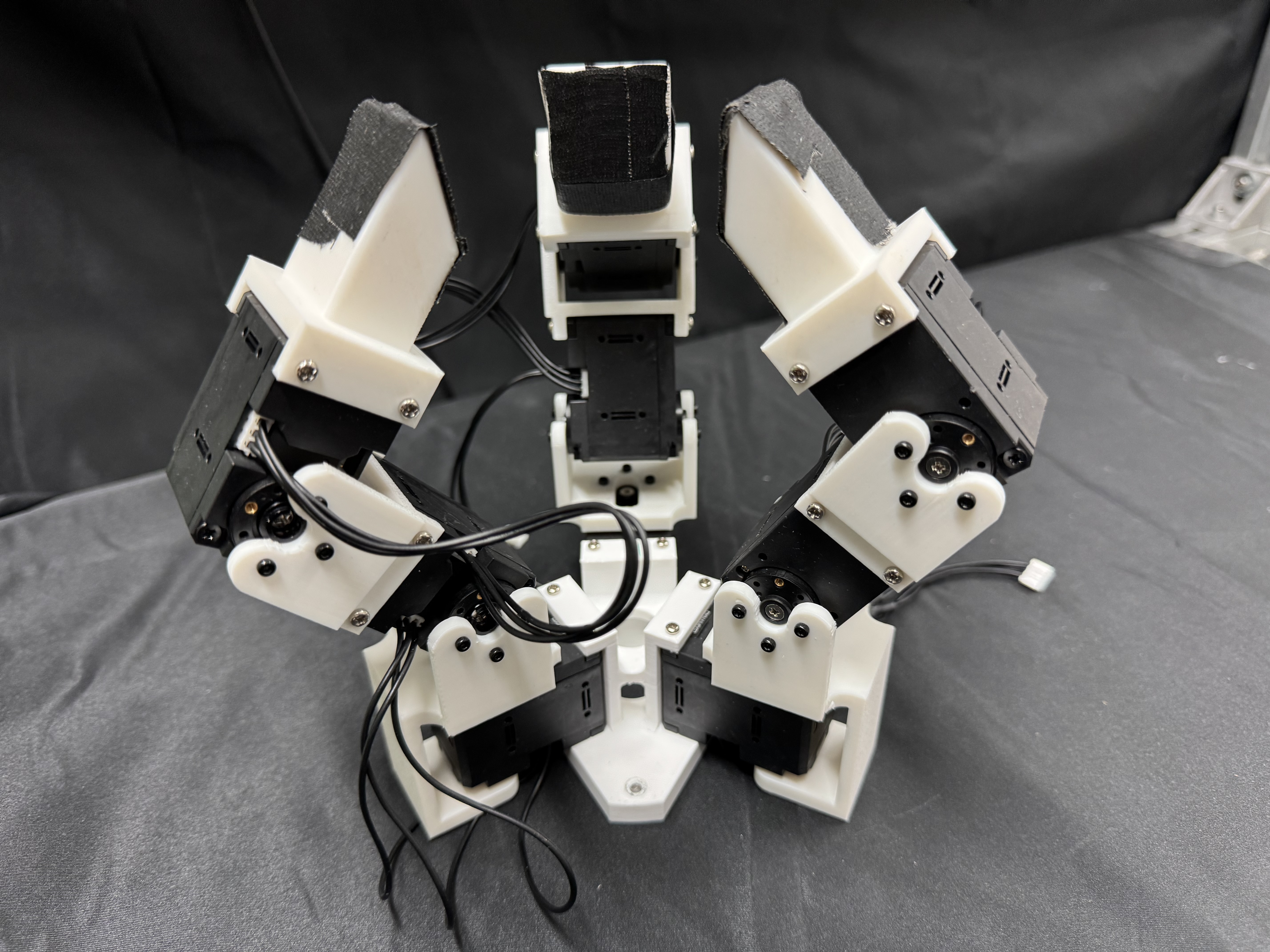} 

    \caption{\textbf{Three-finger Hand Used for the Liquid Handling Task}. 
    \textbf{Left}: The three-finger hand is approaching the pipette.
    \textbf{Right}: The three-finger hand consists of 9 degrees of freedom and uses a flat surface at the fingertips for better gripping performance.
    } 
    \label{fig:three-finger-hand} 
\end{figure}

\begin{figure}[htbp] 
    \centering 
        \includegraphics[width=0.48\textwidth]{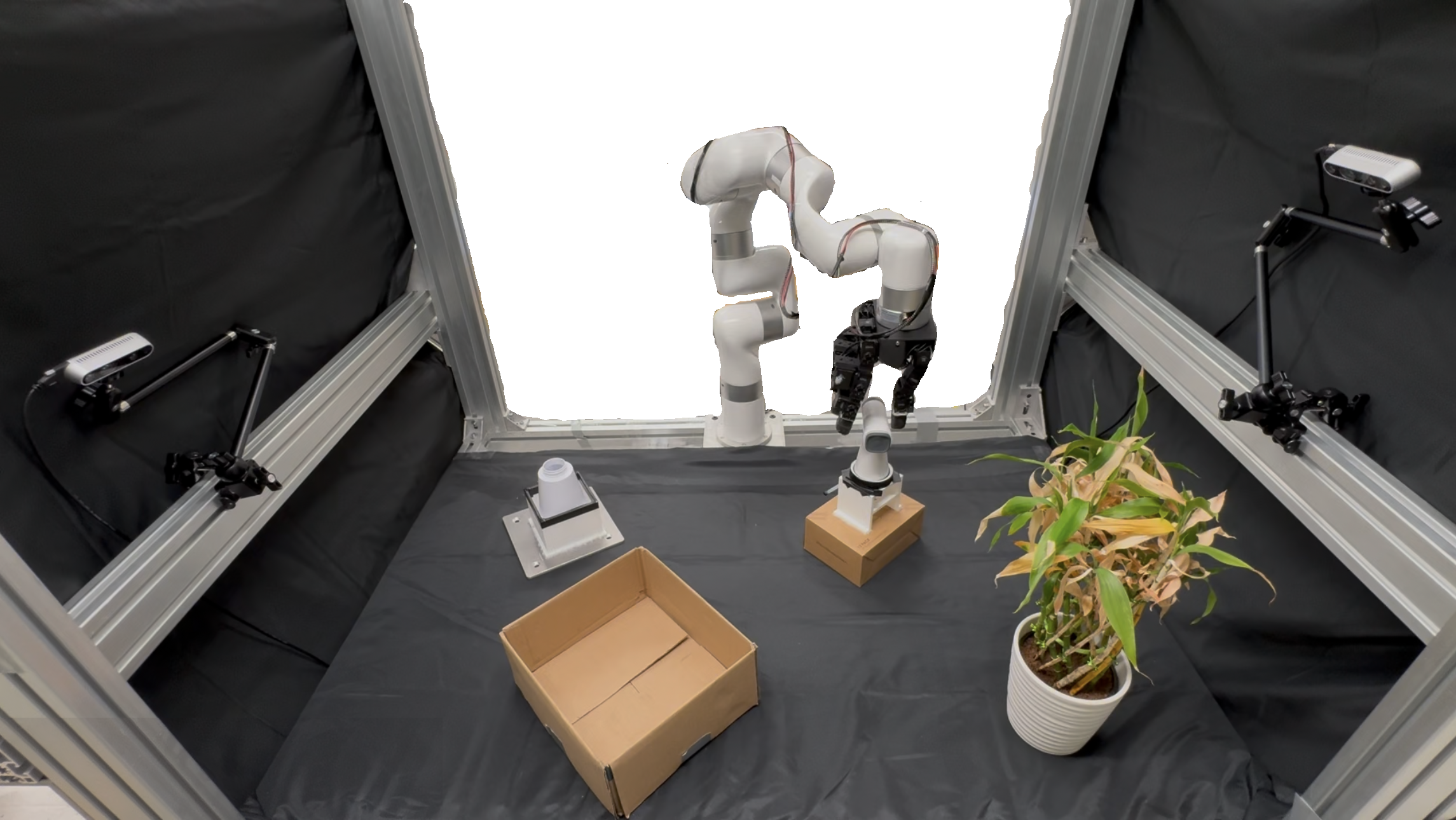} 
        \includegraphics[width=0.48\textwidth]{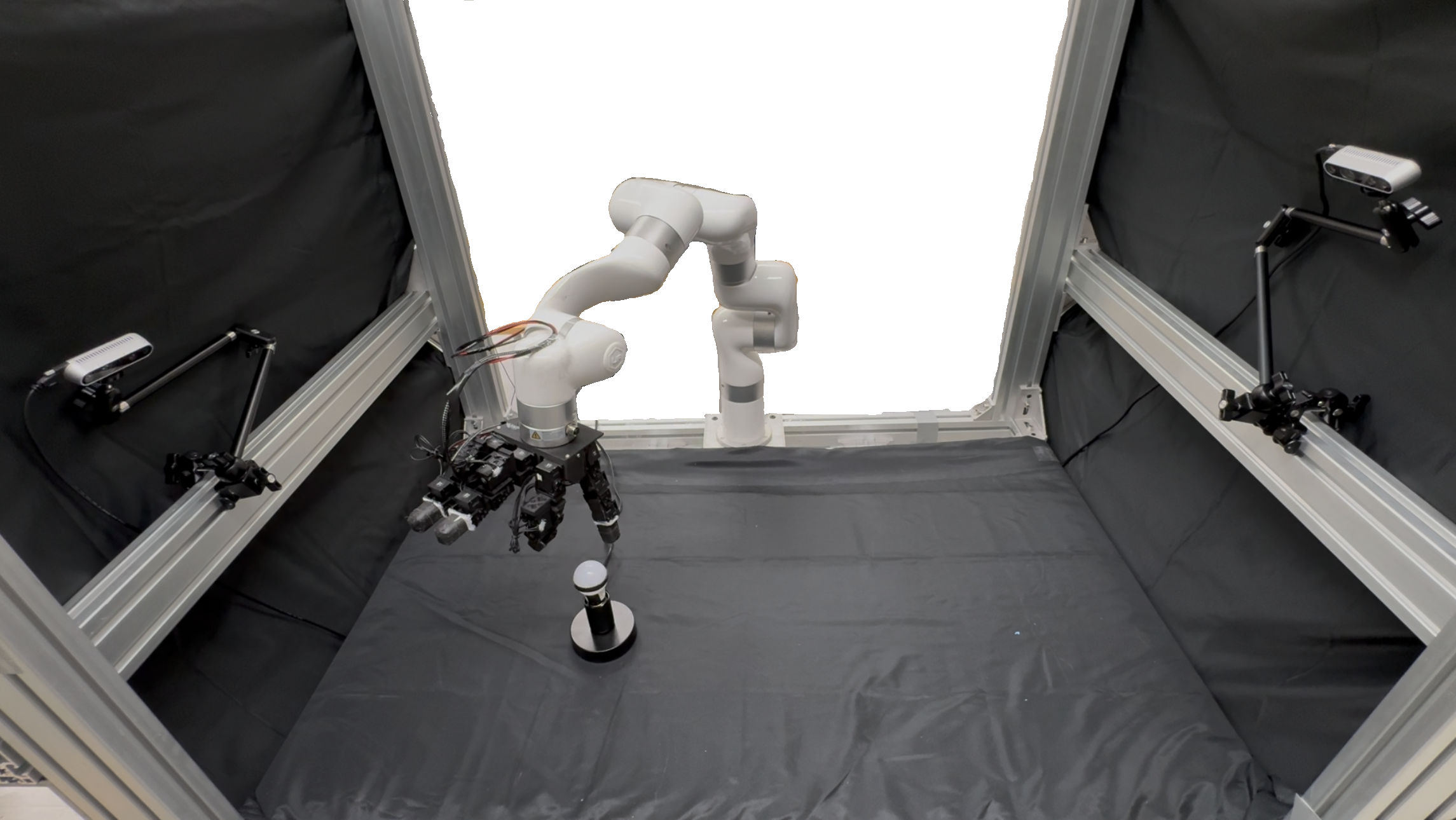} 

    \caption{\textbf{LEAP Hand Used for the Plant Watering and Light Bulb Assembly Tasks}. 
    \textbf{Left}: The LEAP hand is using the thumb, index, and middle fingers to grasp the spray nozzle.
    \textbf{Right}: The LEAP hand is using the thumb and index fingers to screw the bulb in place.
    } 
    \label{fig:leap-hand} 
\end{figure}

For the Liquid Handling task, we use a three-finger hand as shown in Fig.~\ref{fig:three-finger-hand}.
For the Plant Watering and the Light Bulb Assembly tasks, we use a four-finger LEAP hand as shown in Fig.~\ref{fig:leap-hand}.

The three-finger hand is inspired by the DClaw robot~\citep{zhu2019dexterous,ahn2020robel}.
It consists of 9 Dynamixel XL430-W250-T servo motors and has 9 degrees of freedom (DOF) correspondingly.
Compared to the DClaw robot, we replace the metal connectors with 3D printed parts for a lower price and change the finger tips from a sphere to a flat surface for better gripping performance.
The design of the three-finger hand will be open-sourced on the paper website.

We use a light version of the LEAP hand that comprises 16 Dynamixel XL330-M288-T motors and has 16 DOF correspondingly.

We use a joint position controller for both the three-finger hand and the LEAP hand.
The controller receives joint position commands at 20 Hz and sends the commands to the motors at 60 Hz.
Since Dynamixel motors have a PD controller internally, we opt not to do interpolation externally.
Furthermore, friction tape is applied to the fingertips of both hands to enhance friction and improve object grasping.

\subsubsection{Obtaining Point Cloud from Multi-view Cameras}

\begin{figure}[htbp]
    \centering
    \includegraphics[width=0.6\textwidth]{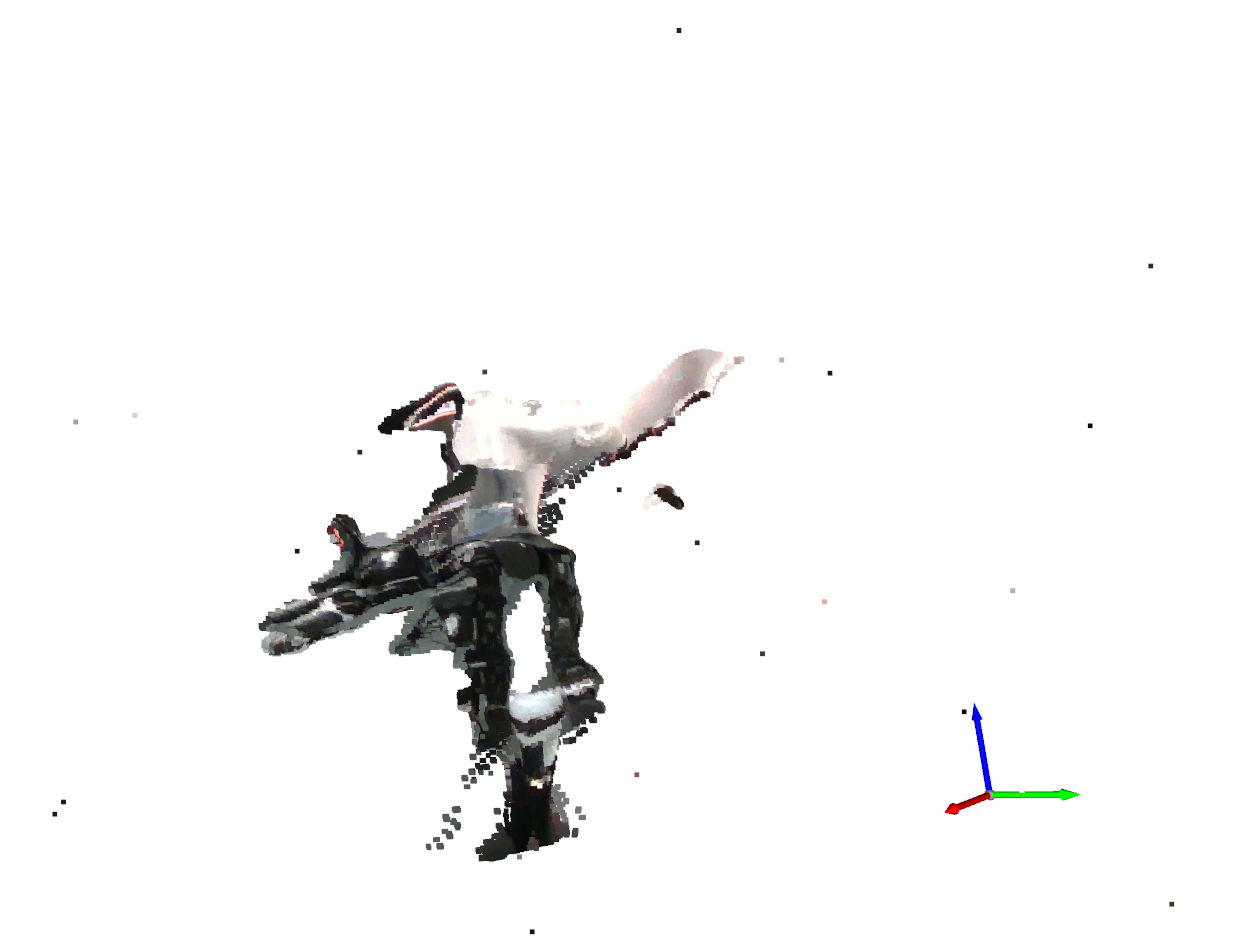}
    \caption{\textbf{Visualization of input real-world point clouds}. 
    The point clouds from the two cameras are concatenated in the robot base coordinate and cropped to be within the workspace.
    We additionally add a Gaussian noise with a small probability to the point clouds to enhance robustness against the sim-to-real gap from sensor noise.
    }
    \label{fig:point-cloud}
\end{figure}

For all three manipulation tasks, we use two RealSense D435 cameras for point cloud reconstruction to address occlusion issues.
In detail, we determine the extrinsics of the cameras in the robot base coordinate by calibrating them with a ChArUco board.
Then, the captured colorized point clouds are transformed into the robot base coordinate and concatenated.
We crop the concatenated point clouds within the robot’s workspace.
Additionally, to address the sim-to-real gap from sensor noise encountered in the real world, we use Flying Point Augmentation~\citep{li2024planning}, where large Gaussian noise is randomly added to the point clouds with 0.5\% probability.
One example is shown in Fig.~\ref{fig:point-cloud}.


\subsection{Teleoperation System}
\label{sec:teleop-system}

\begin{figure}[htbp]
    \centering
    \includegraphics[width=0.5\textwidth]{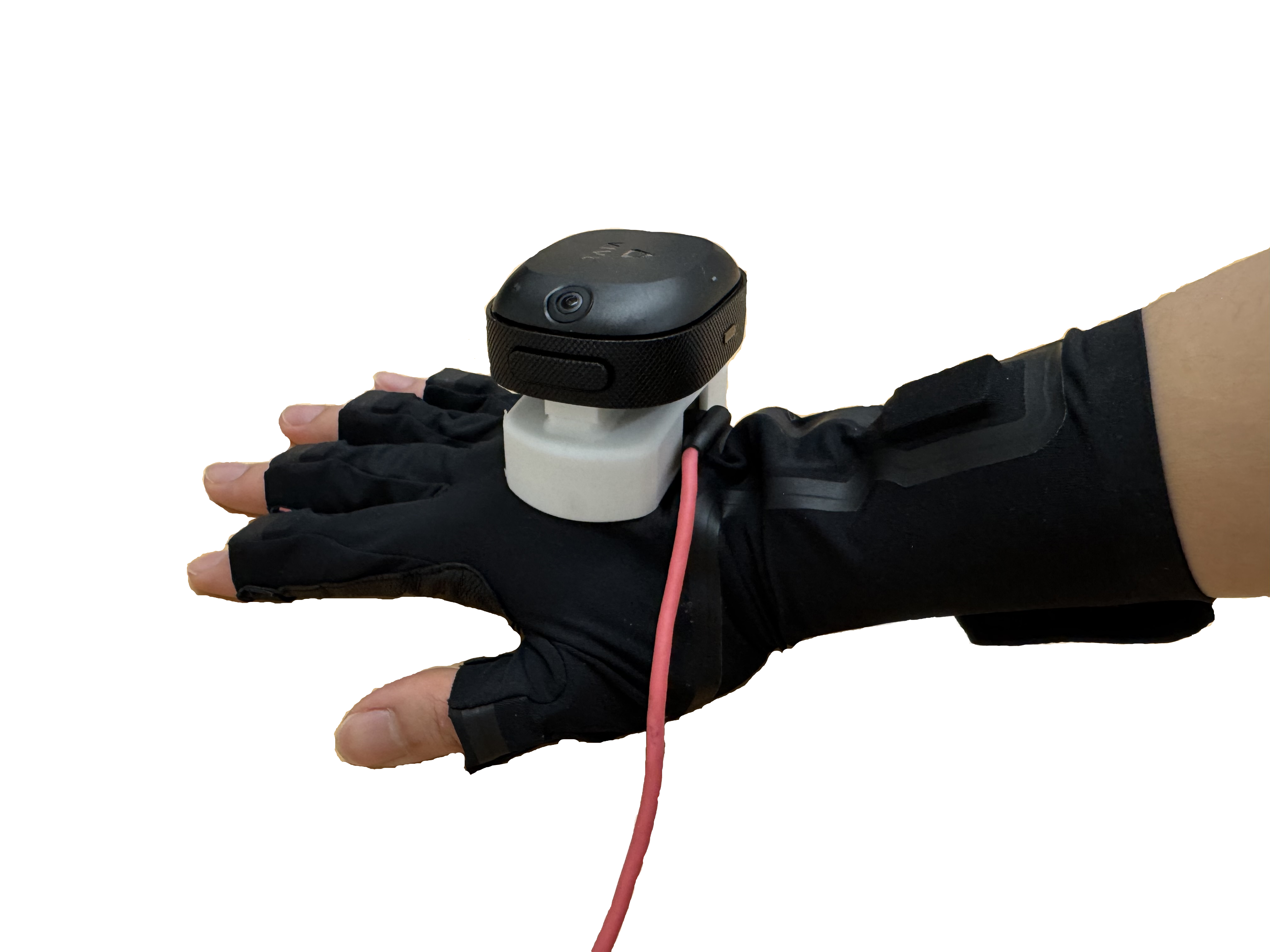}
    \caption{\textbf{Teleoperation System for \ourmethod{}}. 
    We use a Rokoko Smart glove to track finger motions and mount a Vive Ultimate tracker on the glove to track hand poses.
    }
    \label{fig:teleop-system}
\end{figure}

To collect demonstrations for dexterous manipulation requiring high precision and reactivity, the teleoperation system needs to accurately track finger motions and hand poses.
To this end, inspired by DexCap\citep{wang2024dexcap}, we use a Rokoko Smart glove to track finger joint positions and mount a Vive Ultimate tracker on it to estimate hand poses as shown in Fig.~\ref{fig:teleop-system}.
Since the Rokoko glove and the Vive tracker can only stream data to a Windows machine, we forward the tracking data at 50 Hz from the Windows machine to a Ubuntu laptop for controlling the robot.

We apply a low-pass filter to the tracked hand poses and finger joint positions to generate smooth motion for the robot.
We directly use the filtered hand pose from the Vive tracker as the target pose of the end effector of the xArm7 robot.
We calculate the fingertip positions from human data, linearly transform the positions to account for the larger finger sizes of the robotic hands, and use the transformed positions as the target for the fingertips of the robotic hands~\citep{qin2023anyteleop}.
We use mink~\citep{Zakka_Mink_Python_inverse_2024}, a differential inverse kinematics library, to calculate the joint positions for both the robotic hands and the xArm7 robot at the same time.

\subsection{Environment Setup}
\label{sec:environment-setup}

\begin{figure}[htbp] 
    \centering 
        \includegraphics[width=0.32\textwidth]{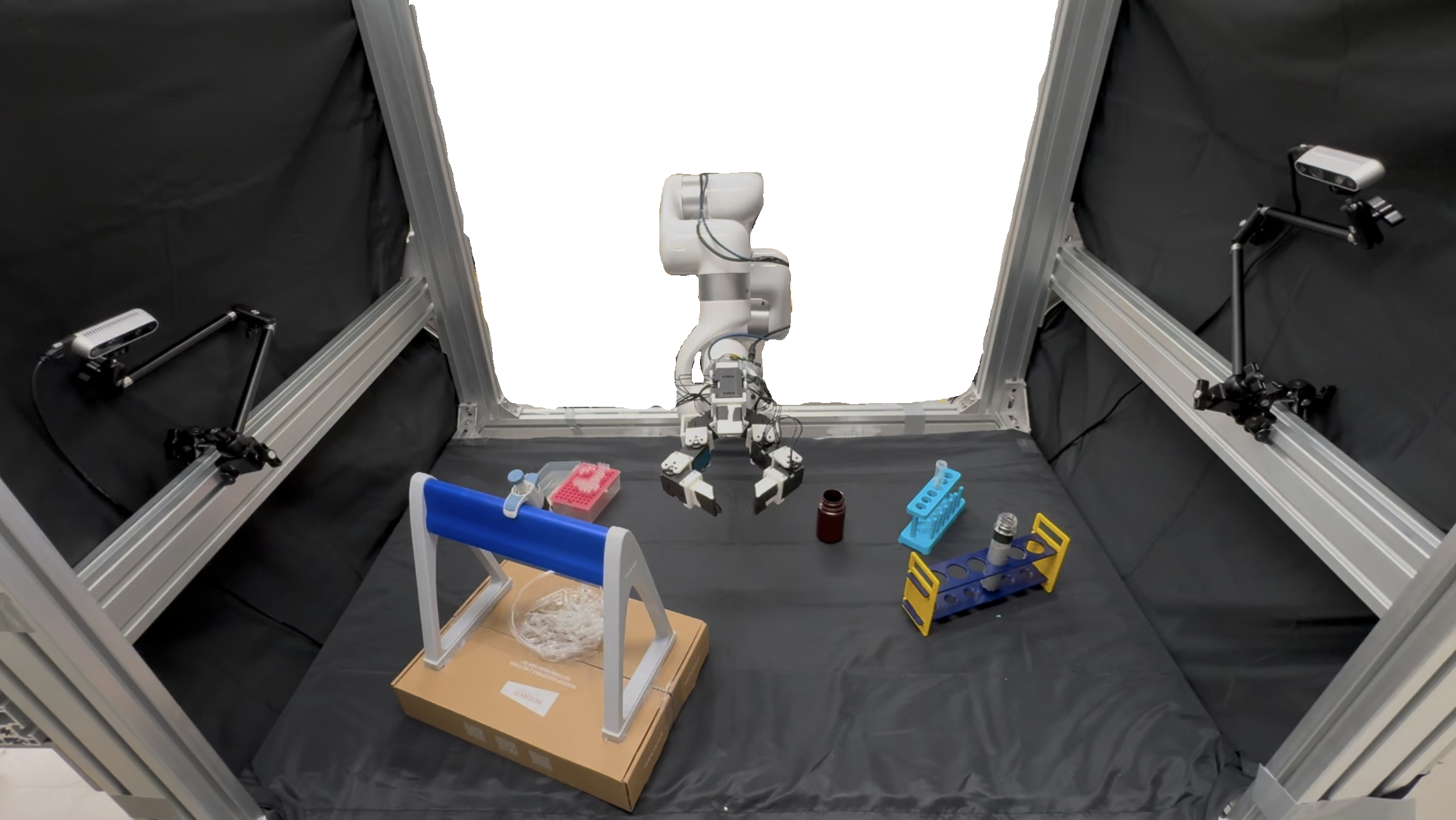} 
        \includegraphics[width=0.32\textwidth]{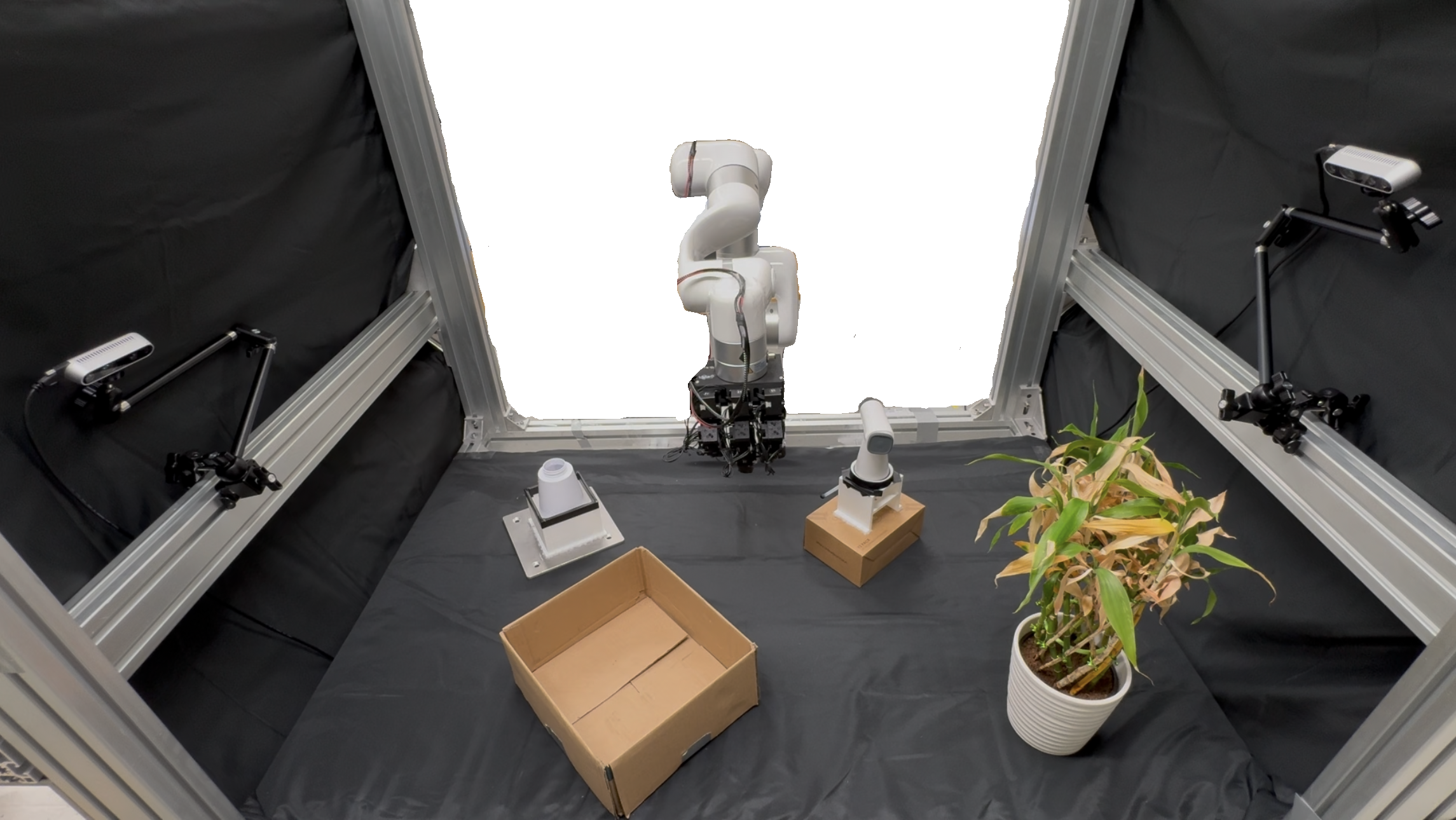} 
        \includegraphics[width=0.32\textwidth]{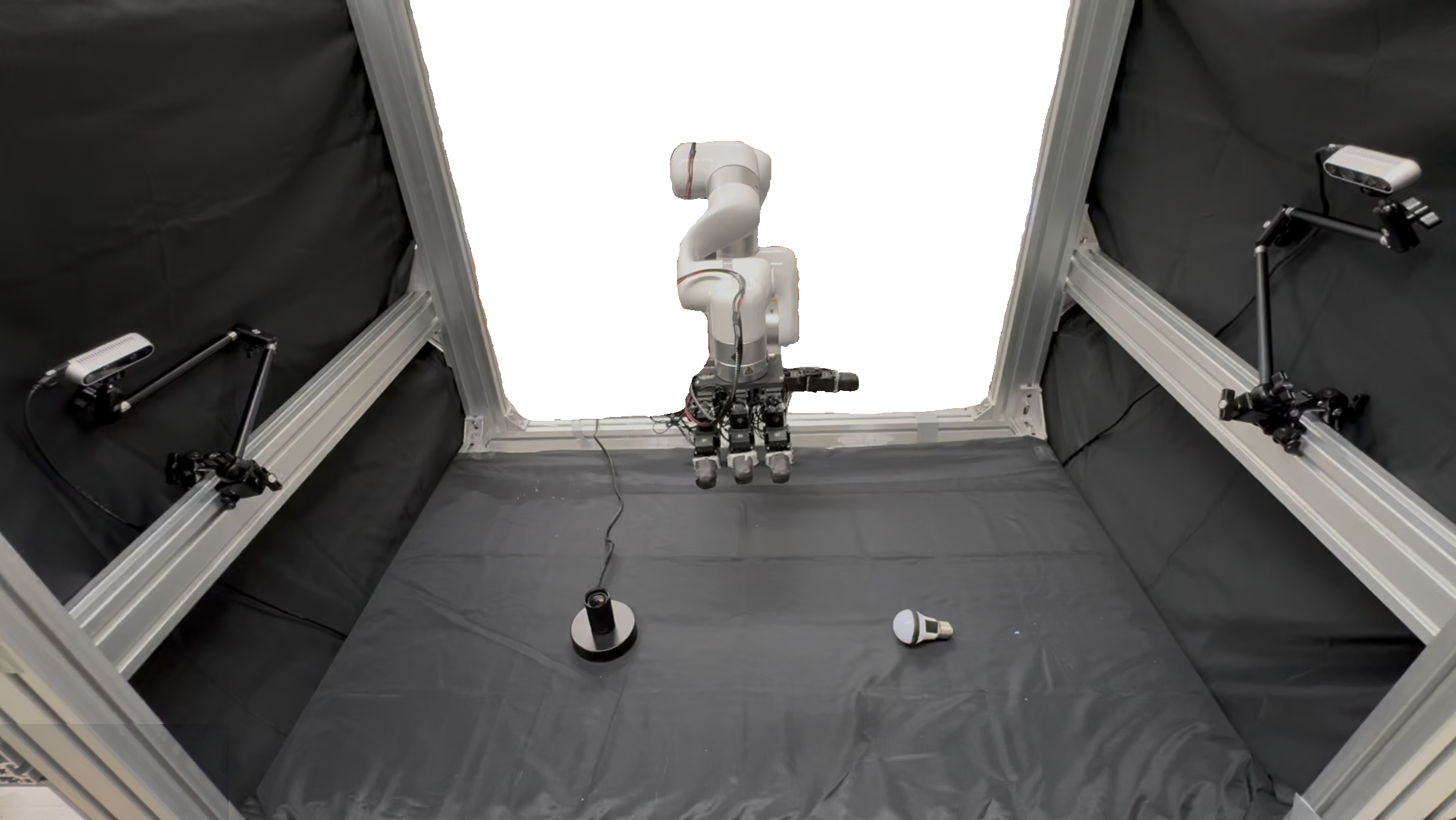} 

    \caption{\textbf{Environment Setup for the three dexterous manipulation tasks}. 
    \textbf{Left}: Liquid Handling.
    \textbf{Middle}: Plant Watering.
    \textbf{Right}: Light Bulb Assembly.
    } 
    \label{fig:real-environment-setup} 
\end{figure}

In this section, we provide the details of the environment setup for the three dexterous manipulation tasks: Liquid Handling, Plant Watering, and Light Bulb Assembly.


\subsubsection{Liquid Handling}
In the Liquid Handling task, the tabletop workspace contains: 1) a pipette, 2) a rack, 3) a reagent bottle, 4) a test tube, 5) a container for dispensed tips, and 6) other objects that commonly appear in a biochemical experiment.
In this task, the robot needs to 
\begin{enumerate}
    \item grasp the pipette from the rack,
    \item aspirate liquid from the reagent bottle,
    \item dispense the liquid into the test tube,
    \item put the pipette back onto the rack, and
    \item dispose of the used tip into the container below.
\end{enumerate}

\subsubsection{Plant Watering}
In the Plant Watering task, there are 1) a spray nozzle, 2) a spray bottle body, 3) a nozzle stand to hold the spray nozzle vertically, 4) a container to hold the bottle body, 5) an open cardboard box, and 6) a potted plant to be watered.
To water the plant, the robot needs to
\begin{enumerate}
    \item grasp the spray nozzle from the nozzle stand,
    \item insert it onto the spray bottle body,
    \item securely twist it in place,
    \item grasp the assembled spray bottle,
    \item press the trigger in front of the plant while holding the bottle, and
    \item place the spray bottle into the cardboard box.
\end{enumerate}

\subsubsection{Light Bulb Assembly}
In the Light Bulb Assembly task, the workspace consists of 1) a light bulb and 2) a bulb base.
To accomplish the task, the robot needs to
\begin{enumerate}
    \item grasp the light bulb,
    \item reorient the bulb in-hand for insertion,
    \item insert the light bulb into the bulb base,
    \item precisely screw the bulb into the base until illumination.
\end{enumerate}

\section{Simulation Training Details}

In this section, we introduce more details about training in simulation, including the used simulator, task designs, skill segmentation, real-to-sim transfer, residual reinforcement learning (residual RL), skill policy training, and skill routing transformer policy.

\subsection{The Simulator}
We use Isaac Gym~\cite{makoviychuk2021isaac} as the simulator backend.
We use the xArm7 model from its ROS package~\citep{githubGitHubXArmDeveloperxarm_ros2} and the LEAP hand model from \citep{shaw2023leaphand}.
For the three-finger hand, we manually designed its 3D model and exported it to URDF.
The 3d models of the objects in the workspace are generated from multi-view images with AR-Code~\citep{arcode2022}. 
Inspired by \citep{torne2024reconciling}, we manually separate the meshes into individual links and define the kinematic relationships for the articulated objects to be manipulated.
See more details in Sec.~\ref{sec:3d-asset-gen}.






\subsection{Skill Segmentation}
We perform frame-level segmentation of long-horizon dexterous demonstrations into multiple skill segments.
This section details the keypoint proposal method and the generation of the discriminator function used for segmenting the demonstration trajectories.

\subsubsection{Keypoint Proposal}
\label{sec:keypoint-proposal}

\begin{figure}[htbp]
    \centering
    \includegraphics[width=1.0\textwidth]{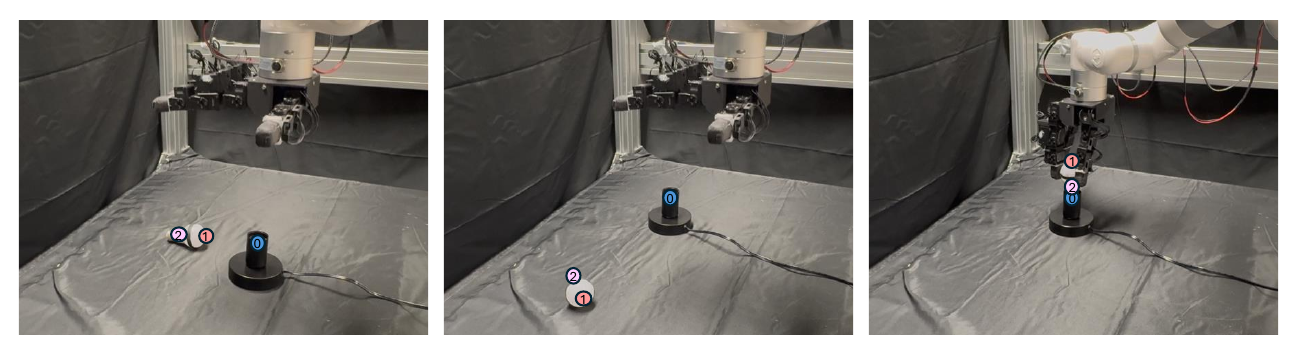}
    \caption{\textbf{Process of keypoint semantic correspondence and tracking.} A semantic correspondence model propagates annotated keypoints from the initial frame of one demonstration (\textbf{left}) to the initial frame of another (\textbf{middle}). Subsequently, a point tracking model tracks these keypoints in further frames of this demonstration (\textbf{right}).
    }
    \label{fig:keypoint-proposal-vis}
\end{figure}

For each of the three demonstration tasks (Liquid Handling, Plant Watering, and Light Bulb Assembly), we randomly select a reference demonstration trajectory.
With this reference trajectory, we manually annotate $N$ semantically meaningful keypoints $\{\hat{p}_l\}_{l=1}^N$ on task-related objects in the first frame.
Then, we propagate these initial keypoints to the first frame of other demonstration trajectories with the semantic correspondence model DIFT~\citep{tang2023emergent}, obtaining consistent keypoints among all the demonstrations.
After that, we track the keypoints for each frame from start to end within each trajectory with Co-Tracker~\citep{karaev2024cotracker}.
A sample for the process above for each task is visualized in Fig.~\ref{fig:keypoint-proposal-vis}.

\subsubsection{Discriminator Function Generation}
We design the language instruction for each task and feed it with the first image and the corresponding tracked keypoints into the OpenAI o3 model, which outputs the total number of skill stages and the corresponding discriminator functions $d_i^{(s)}(s_t)$ in Python.

\subsection{Real-to-Sim Transfer}
Given the segmented demonstration for each skill, we need to create 3D models of the objects and place them at their corresponding positions in the simulation environment.
In this section, we show the details for creating the 3d assets and estimating the object poses.

\subsubsection{3D Asset Generation}
\label{sec:3d-asset-gen}

\begin{figure}[htbp] 
    \centering 
        \includegraphics[width=0.32\textwidth]{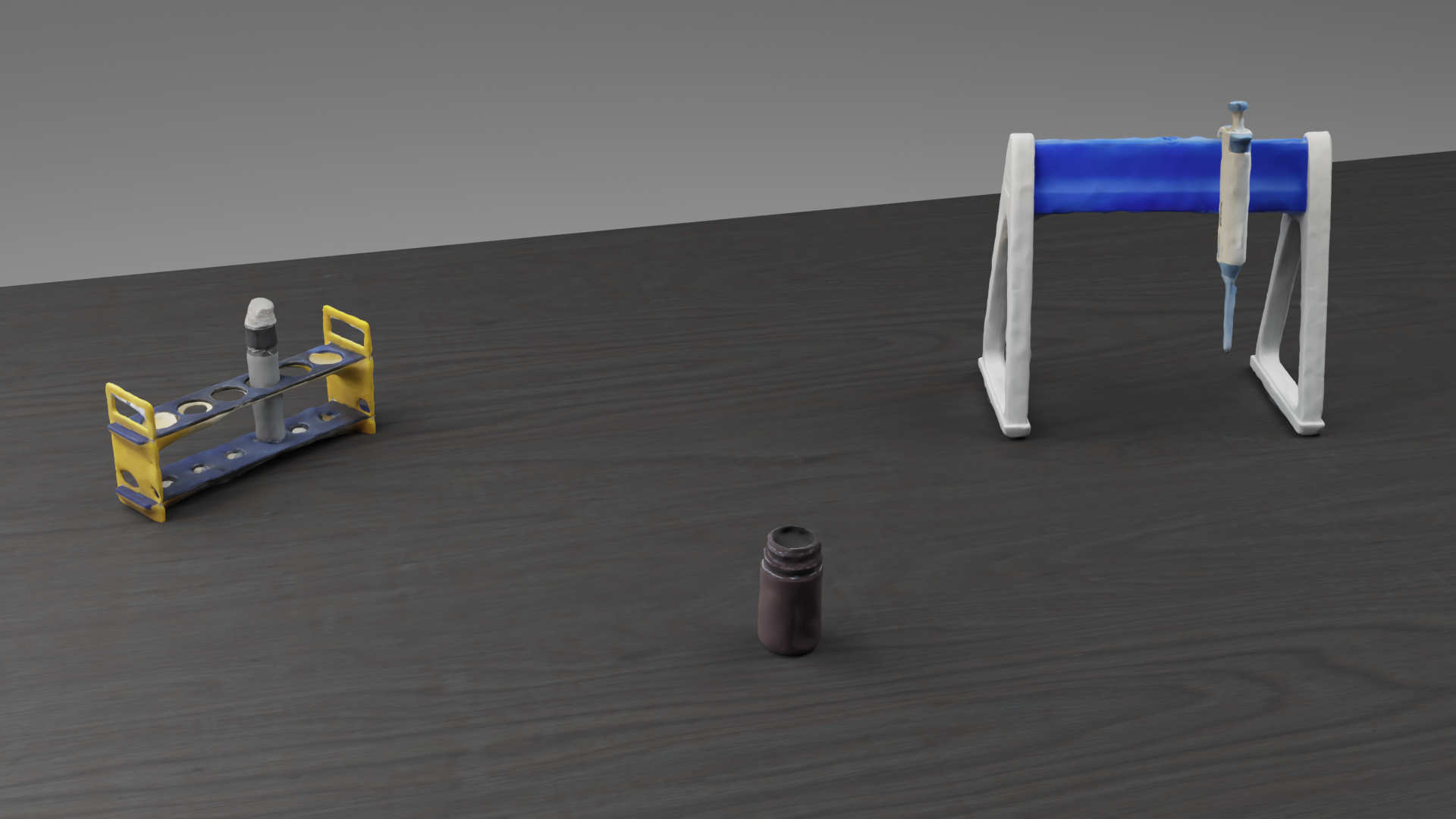} 
        \includegraphics[width=0.32\textwidth]{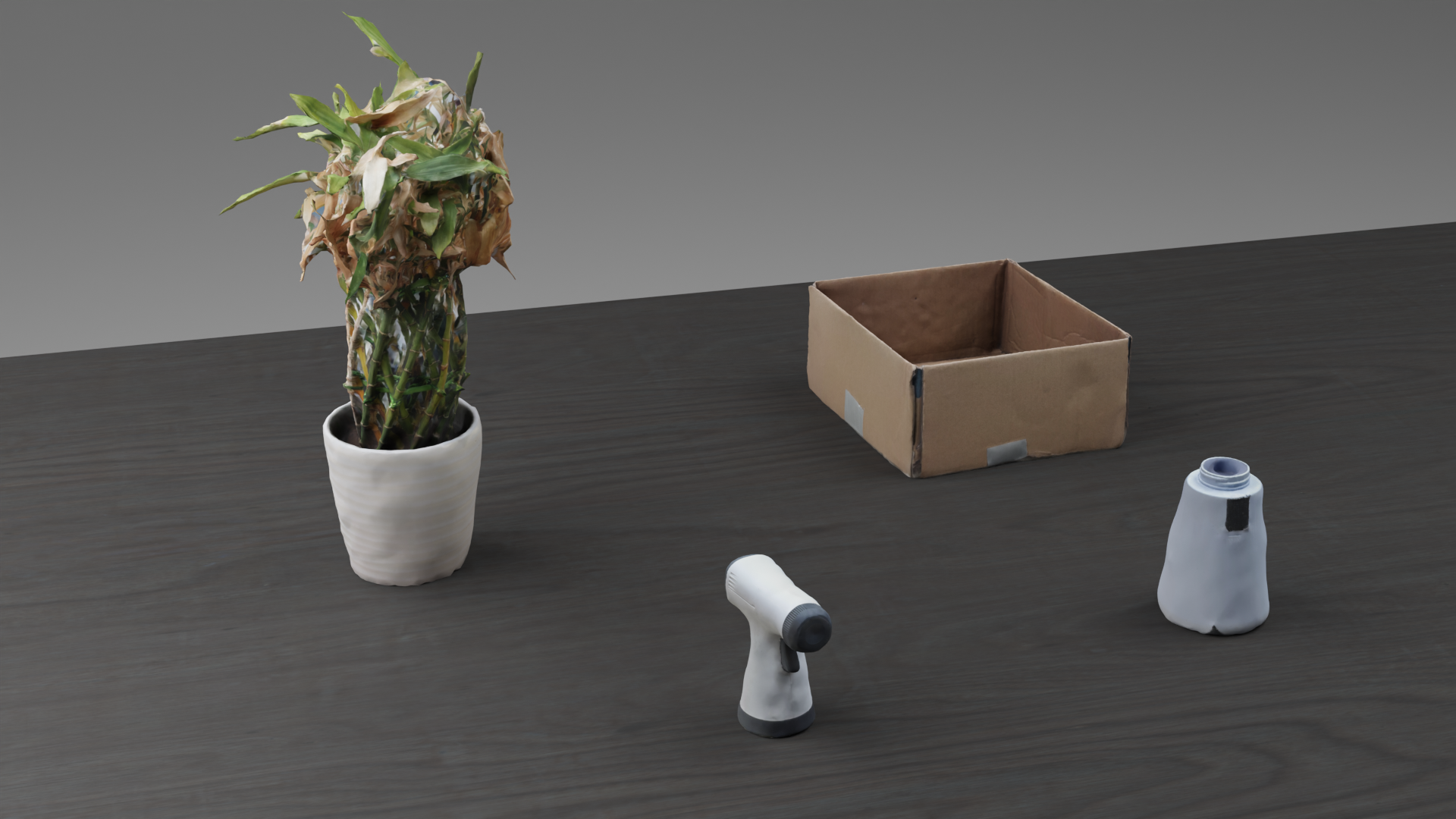} 
        \includegraphics[width=0.32\textwidth]{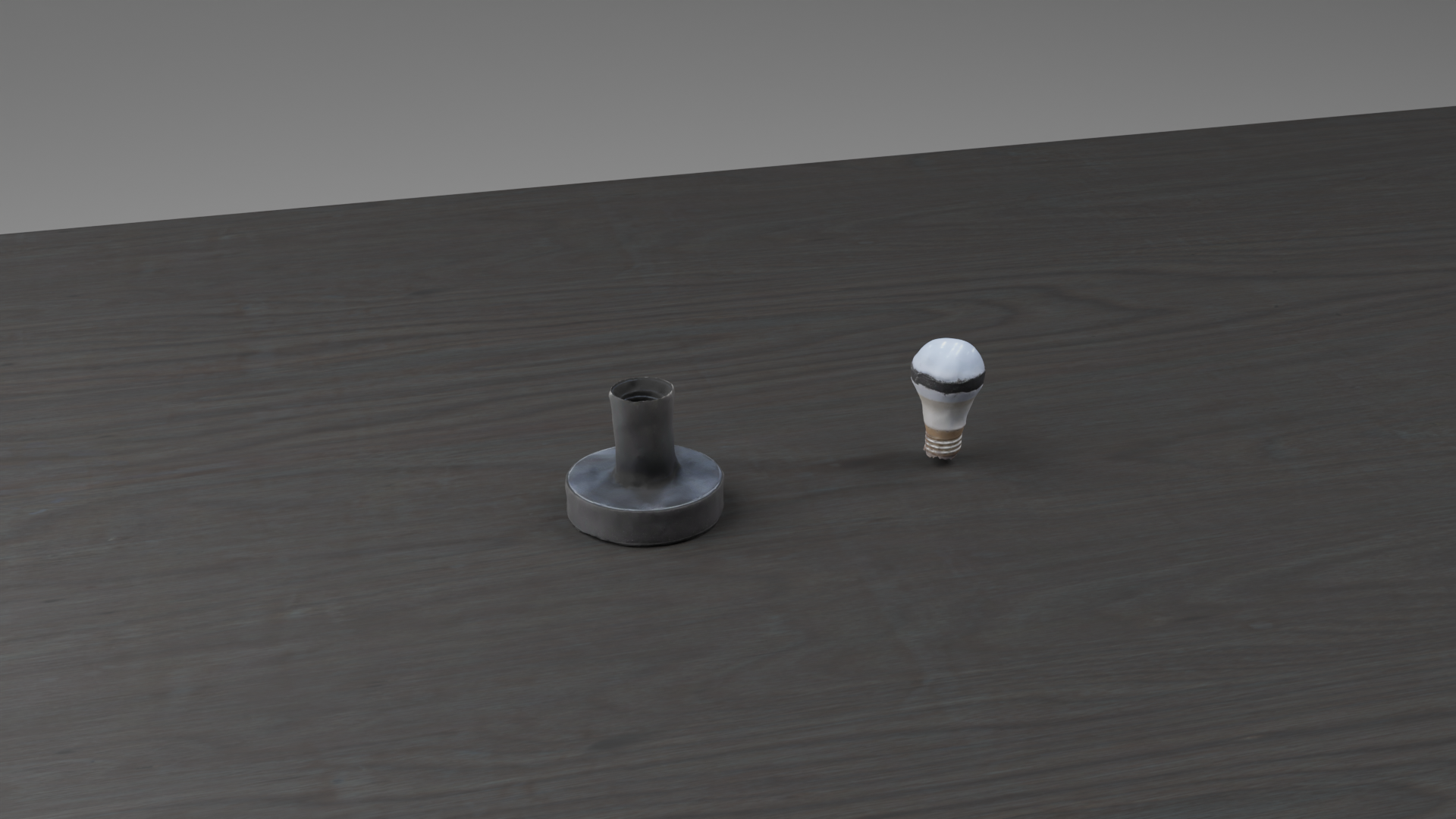} 

    \caption{\textbf{Textured Meshes Created for the three manipulation tasks}.
    \textbf{Left}: Liquid Handling.
    \textbf{Middle}: Plant Watering.
    \textbf{Right}: Light Bulb Assembly.
    } 
    \label{fig:mesh-vis} 
\end{figure}
We use AR-Code~\citep{arcode2022} to create the textured mesh for each object by capturing multi-view images. This will take around 3-5 minutes for each object.
We show a full list of the meshes created in Fig.~\ref{fig:mesh-vis}.
Following the methodology in \citep{torne2024reconciling} for articulated objects, we manually decompose their meshes into individual links and establish their kinematic relationships by defining articulations which take around 3-5 minutes for each articulated objects.



\subsubsection{State Estimation}

Accurate estimation of object poses within the workspace is crucial for seamlessly transferring objects to simulation and generating synthetic augmented trajectories for skill training. 
We leverage tracked keypoints (see Sec.~\ref{sec:keypoint-proposal}) to prompt SAM2~\citep{ravi2024sam2} and obtain a segmentation mask for each object of interest. 
Using this segmentation mask and the reconstructed 3D mesh, we then employ FoundationPose~\citep{wen2024foundationpose} to estimate the objects' 6D poses.


\subsubsection{Domain Randomization}

To enhance the robustness against the sim-to-real gap for each skill policy, we apply domain randomization to various simulation properties for the objects involved.
The randomization parameters are given in Tab.~\ref{tab:domain_randomization}.

\begin{table}[htbp]
\centering
\begin{tabular}{lll@{\hskip 20pt}lll}
\toprule
\textbf{Parameter} & \textbf{Type} & \textbf{Distribution} & \textbf{Parameter} & \textbf{Type} & \textbf{Distribution} \\
\midrule
object mass              & Scaling & $\mathcal{U}(0.5,\ 1.5)$       & joint lower range & Additive & $\mathcal{N}(0,\ 0.01)$ \\
object static friction          & Scaling & $\mathcal{U}(0.7,\ 1.3)$       & joint upper range & Additive & $\mathcal{N}(0,\ 0.01)$ \\
robot static friction & Scaling & $\mathcal{U}(0.7,\ 1.3)$ & joint damping     & Scaling & $\mathcal{E}(0.3,\ 3.0)$ \\
state observation & Additive & $\mathcal{U}(-0.002,\ 0.002)$ &  joint stiffness     & Scaling & $\mathcal{E}(0.75,\ 1.5)$ \\
action & Additive & $\mathcal{N}(0,\ 0.01)$ & gravity     & Scaling & $\mathcal{U}(0.9,\ 1.1)$ \\
restitution & Scaling & $\mathcal{U}(0.5,\ 1.5)$ & Compliance     & Scaling & $\mathcal{U}(0.5,\ 1.5)$ \\
\bottomrule
\end{tabular}
\vspace{2mm}
\caption{
\textbf{Randomization ranges for physical and control parameters.} 
$\mathcal{U}(a, b)$ denotes uniform distribution over $[a, b]$. 
$\mathcal{N}(\mu, \sigma^2)$ denotes Gaussian distribution with mean $\mu$ and variance $\sigma^2$. 
$\mathcal{E}(a, b)$ denotes exponential-uniform distribution, i.e., $\mathcal{E}(a, b) = \exp(\mathcal{U}(\log a, \log b))$.
}
\label{tab:domain_randomization}
\end{table}

\subsection{Residual Reinforcement Learning}
Given the converted skill segments in simulation, we generate synthetic augmentation data to improve the robustness and generalization for each skill policy.
We utilize residual reinforcement learning (residual RL) to create diverse simulated demonstrations.
In this section, we introduce the details for the base policy and the residual policy in residual RL.

\subsubsection{Base Policy Training}
With the skill segments converted into simulation, we first train a state-based base policy using Behavior Cloning. This base policy utilizes a similar diffusion-based architecture in~\citep{wang2024rise} (the vision encoder is replaced with an MLP to encoder object pose) and outputs an action chunk \citep{zhao2023learning,chi2023diffusion} to predict a set of future actions.
The observation space of the base policy is shown in Tab.~\ref{tab:obs_base} .The detailed parameters for training the base policy are given in Tab.~\ref{tab:diffusion_hyperparams}.


\begin{table}[h]
\centering
\begin{tabular}{ll@{\hskip 20pt}ll}
\toprule
\textbf{Name} & \textbf{Dimension} & \textbf{Name} & \textbf{Dimension} \\
\midrule
Arm joint position  & 7 & Hand joint position & 9 / 16 \\
Arm joint velocity  & 7 & Hand joint velocity & 9 / 16 \\
Object Pose  & $N_{obj}$ * 6 & Fingertip pose & $N_{finger}$ * 6 \\
\bottomrule
\end{tabular}
\vspace{2mm}
\caption{\textbf{The observation space of base policy.}}
\label{tab:obs_base}
\end{table}

\begin{table}[h]
\centering
\begin{tabular}{ll@{\hskip 20pt}ll}
\toprule
\textbf{Parameter} & \textbf{Value} & \textbf{Parameter} & \textbf{Value} \\
\midrule
Downsampling dimensions & [256, 512, 1024] & Embedding dimensions & 64 \\
Kernel size & 5& Observation horizon & 2 \\
Prediction horizon & 16 & Action horizon & 8 \\
DDPM training steps & 100 & DDIM inference steps & 4 \\
Gradient steps & 200000 & Batch size & 1024 \\
Learning rate & 1e-4 & Optimizer & AdamW \\
\bottomrule
\end{tabular}
\vspace{2mm}
\caption{\textbf{State-based diffusion policy hyperparameters.}}
\label{tab:diffusion_hyperparams}
\end{table}

\subsubsection{Residual Policy Training}
We utilize Proximal Policy Optimization (PPO), a model-free RL algorithm, to learn a residual policy complementing the base policy. 
The residual policy use a simple MLP architecture.
The observation space of residual policy is shown in Tab~\ref{tab:obs_res}.
The hyperparameters used in PPO training are given in Tab.~\ref{tab:ppo_hyperparams}.
In addition, we use orthogonal initialization~\citep{saxe2013exact} and progressive exploration schedule~\citep{yuan2024policy} for the residual policy learning.
The progressive exploration schedule employs an $\epsilon$-greedy strategy where, at each step, the agent uses the residual policy with probability $\epsilon$ and otherwise relies solely on the base policy. 
The probability $\epsilon$ increases linearly from 0 to 1 over $H$ environment steps, where $H = 100\mathrm{K}$.




\begin{table}[h]
\centering
\begin{tabular}{ll@{\hskip 20pt}ll}
\toprule
\textbf{Name} & \textbf{Dimension} & \textbf{Name} & \textbf{Dimension} \\
\midrule
Arm joint position  & 7 & Hand joint position & 9 / 16 \\
Arm joint velocity  & 7 & Hand joint velocity & 9 / 16 \\
Object states  & $N_{obj}$ * 13 & fingertip state & $N_{finger}$ * 13 \\
Contact Forces & $N_{obj}$ * 3 & - & - \\
\bottomrule
\end{tabular}
\vspace{2mm}
\caption{\textbf{The observation space of residual policy.} The object and fingertip state includes position, orientation, linear velocities and angular velocities.}
\label{tab:obs_res}
\end{table}

\begin{table}[h]
\centering
\begin{tabular}{ll@{\hskip 20pt}ll}
\toprule
\textbf{Hyperparameter} & \textbf{Value} & \textbf{Hyperparameter} & \textbf{Value} \\
\midrule
PPO rollout steps & 8 & Batches per agent & 4 \\
Learning epochs & 5 & Desired KL & 0.16 \\
Episode length & 200 & Policy iteration threshold & 0.005 \\
Discount factor & 0.96 & GAE parameter & 0.95 \\
Entropy coeff. & 0.0 & PPO clip range & 0.2 \\
Learning rate & 0.0003 & Value loss coeff. & 1.0 \\
Max gradient norm & 1.0 & Initial noise std. & 0.8 \\
Clip actions & 1.0 & Clip observations & 5.0 \\
\bottomrule
\end{tabular}
\vspace{2mm}
\caption{\textbf{Hyperparameters for RL policy.}}
\label{tab:ppo_hyperparams}
\end{table}

\subsection{Skill Policy Training}
We use 1000 generated simulation demonstrations and 15 real world demonstrations to co-train the final skill policy. The observation space of the skill policy is shown in Tab.~\ref{tab:obs_skill}. We adopt a diffusion-based policy architecture similar to that of the base policy.
Specifically, for \ourmethodpc{}, we utilize the policy architecture from~\cite{wang2024rise}, which comprises a 3D encoder built with sparse convolutions for point cloud encoding and a diffusion-based action head.
For \ourmethodpose{}, this 3D sparse encoder is replaced with an MLP.

\begin{table}[h]
\centering
\begin{tabular}{ll@{\hskip 20pt}ll}
\toprule
\textbf{Name} & \textbf{Dimension} & \textbf{Name} & \textbf{Dimension} \\
\midrule
Arm joint position  & 7 & Hand joint position & 9 / 16 \\
Arm joint velocity  & 7 & Hand joint velocity & 9 / 16 \\
Initial object pose  & $N_{obj}$ * 6 & Point num & 1024  \\
\bottomrule
\end{tabular}
\vspace{2mm}
\caption{\textbf{The observation space of skill policy.} The initial object pose is used for \ourmethodpose{} and the point cloud is used for \ourmethodpc{}.}
\label{tab:obs_skill}
\end{table}

\subsection{Skill Routing Transformer (SRT) Policy}
Given the trained skill policies, we train a Skill Routing Transformer (SRT) policy to chain them to accomplish the long-horizon task.
In this section, we give the details for generating the transition dataset and policy training.

\subsubsection{Transition Data Generation}
For generating transition trajectories between skill segments, we randomly sample state pairs $(s^\text{end},s^\text{start})\sim \mathcal{E}_{i-1}^\text{aug}\times \mathcal{I}_{i}^\text{aug}$, where $\mathcal{E}_{i-1}^\text{aug}$ is the termination set for the previous skill segment and $ \mathcal{I}_{i}^\text{aug}$ is the initial set for the subsequent skill segment.
For each sampled state pair, we then utilize cuRobo~\citep{sundaralingam2023curobo} to generate a collision-free transition trajectory.
We continue this process until 1000 trajectories have been successfully generated for each specific transition motion between consecutive skills.



\subsubsection{Policy Training}
\begin{figure}[htbp]
    \centering
    \includegraphics[width=1.0\textwidth]{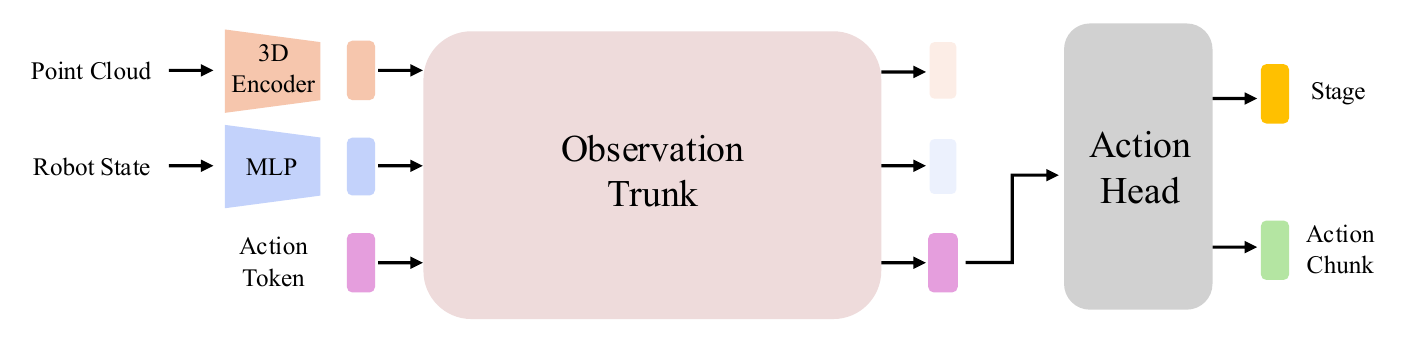}
    \caption{\textbf{Skill Routing Transformer Policy Architecture}. 
    }
    \label{fig:SRT}
\end{figure}

For Skill Routing Transformer (SRT) policy, we use similar architecture with BAKU~\citep{haldar2024baku}, an efficient transformer for multi-task policy learning. The SRT policy architecture is shown in Fig.~\ref{fig:SRT} .
We use a 3D encoder built with sparse convolution~\citep{wang2024rise} to encode point cloud input and an MLP to encode the robot state. The observation trunk use minGPT~\citep{karpathy2021mingpt} architecture and the action head use an MLP.
The action head output an action chunk and stage (to decide either a skill or transition).
The parameters of SRT policy is shown in Tab.~\ref{tab:srt_hyperparams}.
The SRT policy utilizes the same point cloud observation space as the skill policy that takes point cloud as input, with this observation space detailed in Tab.~\ref{tab:obs_skill}.

\begin{table}[h]
\centering
\begin{tabular}{ll@{\hskip 20pt}ll}
\toprule
\textbf{Hyperparameter} & \textbf{Value} & \textbf{Hyperparameter} & \textbf{Value} \\
\midrule
Observation trunk & Transformer & Transformer architecture & minGPT~\citep{karpathy2021mingpt} \\
Transformer hidden dim & 256 & Observation history length & 10 \\
Voxel size (mm) & 5 & Point feature dim & 512 \\
Encoding block & 4 & Decoding block & 1 \\
Action head & MLP & Action head hidden dim & 512 \\
Action head hidden ddepth & 2 & Action chunking length & 10 \\
Optimizer & Adam & Mini-batch size & 64 \\
\bottomrule
\end{tabular}
\vspace{2mm}
\caption{\textbf{Hyperparameters for Skill Routing Transformer policy.}}
\label{tab:srt_hyperparams}
\end{table}

\section{Experiment Settings and Evaluation Details}

\subsection{Training Demonstration Dataset}

By using the teleoperation system introduced in Sec.~\ref{sec:teleop-system}, we collect 15 trajectories for each of the three manipulation tasks (Liquid Handling, Plant Watering, and Light Bulb Assembly).

\subsection{Experiment Settings}
For each task, we use 15 real trajectories to train \ourmethod{} and all the baselines (Real-only, T-STAR~\citep{lee2021adversarial}, Seq-Dex~\citep{chen2023sequential}, MimicGen~\citep{mandlekar2023mimicgen}, and SkillMimicGen~\citep{garrett2024skillmimicgen}).

We evaluate \ourmethod{} and all the baselines for 20 trials.
We start each trial with different poses for both the robot and the objects in the workspace.
We try our best to ensure consistent initial conditions for the evaluation of different methods.
For each of the three manipulation tasks, the robot has to sequentially complete the operations defined in Sec.~\ref{sec:environment-setup}.
After completing the operations sequentially, we terminate each trial and call it a \textit{success} when
\begin{itemize}
    \item in the Liquid Handling task, the tip falls into the container below,
    \item in the Plant Watering task, the spray bottle is put into the cardboard box,
    \item in the Light Bulb Assembly task, the light bulb is illuminated.
\end{itemize}


\subsection{Robustness under out-of-distribution (OOD) conditions}
\begin{figure}[htbp]
    \centering
    \includegraphics[width=1.0\textwidth]{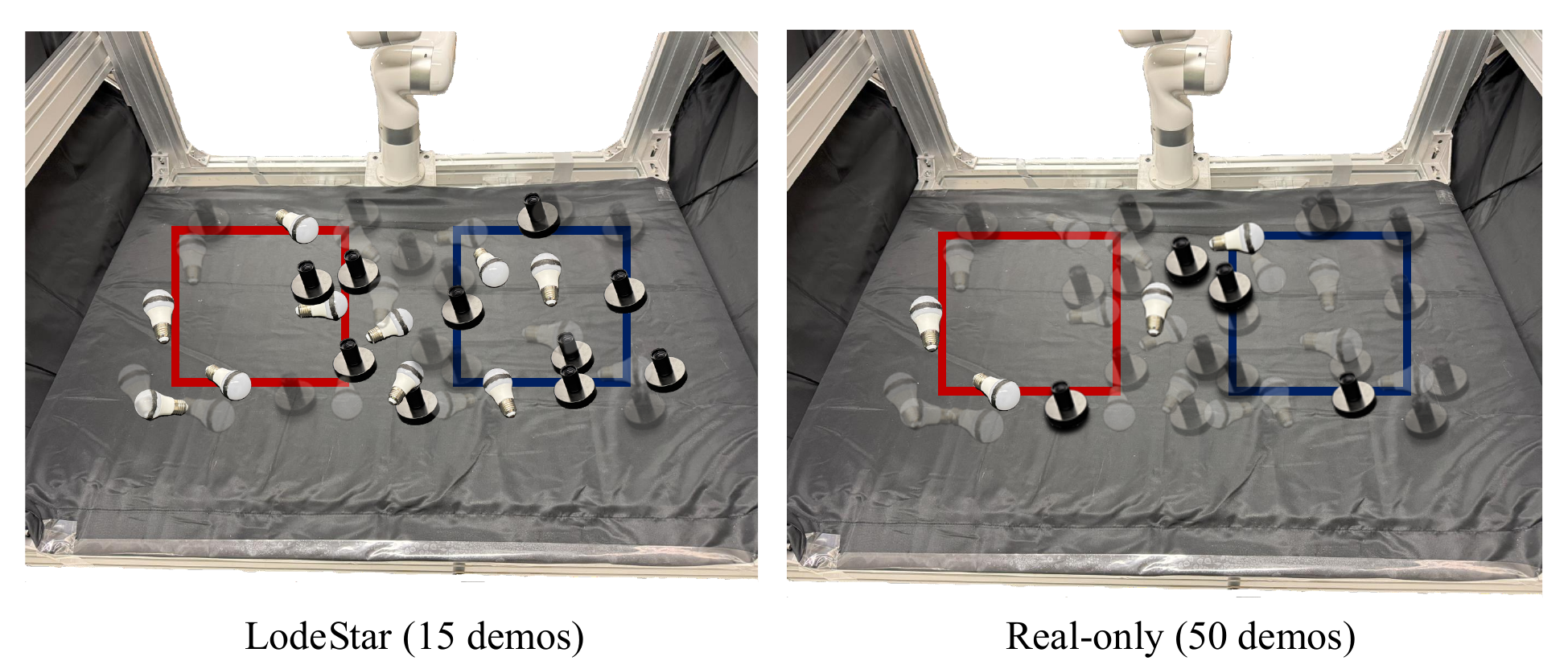}
    \caption{\textbf{OOD Evaluation on Larger Init Distributions}. 
    }
    \label{fig:ood}
\end{figure}

We visualize the OOD experiments with a larger initial distribution condition in Fig~\ref{fig:ood}. The red and blue rectangles represent the distribution of demonstration data for the initial pose of the light bulb and the black base. The multiple light bulbs and black bases illustrate different initial positions across evaluation episodes. Faded ones indicate failures, while solid ones indicate successes. Lodestar demonstrates stronger generalizability than the real-only baseline with fewer demonstrations.

\subsection{Failure Cases Analysis}
\begin{figure}[htbp]
    \centering
    \includegraphics[width=1.0\textwidth]{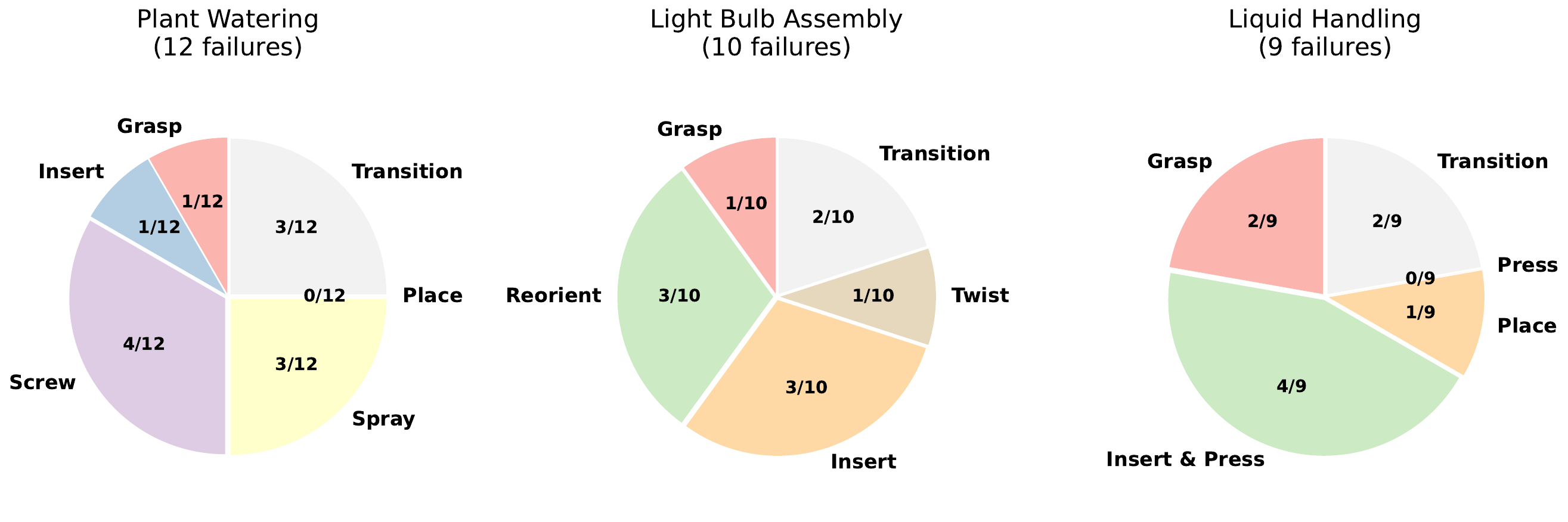}
    \caption{\textbf{Failure Cases Distribution on Three Tasks}. 
    }
    \label{fig:failure_cases}
\end{figure}

We performed a statistical analysis of the failure occurrences of \ourmethod{} during the main experiments, with the specific stages of failure illustrated in Fig.\ref{fig:failure_cases}. The data indicate that a majority of failures manifested across various skill stages. However, a subset of failures also occurred during the transition stages. Our investigation revealed that these transition stage failures are typically attributable to two primary causes: (1) even in instances of passive contact, minute in-hand displacements occasionally led to the object falling; and (2) occasional collisions where the grasped object, undergoing rapid motion during the transition phase, inadvertently struck another object immediately prior to the intended contact.

\end{document}